\documentclass[letterpaper]{article} 
\usepackage{aaai24}  
\usepackage{times}  
\usepackage{helvet}  
\usepackage{courier}  
\usepackage[hyphens]{url}  
\usepackage{graphicx} 
\usepackage{natbib}  
\usepackage{caption} 
\usepackage{xcolor}
\usepackage{multirow}
\usepackage{subfigure}
\usepackage{CJKutf8}
\usepackage[utf8]{inputenc} 

\usepackage{algorithm}
\usepackage{algorithmic}
\usepackage{amsfonts}
\usepackage{amsmath}
\usepackage{bbding}

\usepackage{newfloat}
\usepackage{listings}
\DeclareCaptionStyle{ruled}{labelfont=normalfont,labelsep=colon,strut=off} 
\floatstyle{ruled}
\newfloat{listing}{tb}{lst}{}
\floatname{listing}{Listing}
\title{GlyphDraw2: Automatic Generation of Complex Glyph Posters with Diffusion Models and Large Language Models}

\author{
    Jian Ma\textsuperscript{\rm 1},
    Yonglin Deng\textsuperscript{\rm 2}\thanks{The author did his work during internship at OPPO AI Center.}, 
    Chen Chen\textsuperscript{\rm 1 \#}, 
    Nanyang Du\textsuperscript{\rm 3}\thanks{The author did his work during internship at OPPO AI Center. \# denotes corresponding authors.}, 
    Haonan Lu\textsuperscript{\rm 1}, 
    Zhenyu Yang\textsuperscript{\rm 1} 
}
\affiliations{
    \textsuperscript{\rm 1}OPPO AI Center\\ 
    \textsuperscript{\rm 2}The Chinese University of Hong Kong, Shenzhen\\
    \textsuperscript{\rm 3}Tsinghua University\\

    majian2@oppo.com, yonglindeng@link.cuhk.edu.cn, chenchen4@oppo.com, dny22@mails.tsinghua.edu.cn, \{luhaonan,yangzhenyu\}@oppo.com\\
}

\begin{document}
\maketitle

\begin{figure*}[tb]
	\centering
    \includegraphics[width=1\textwidth]{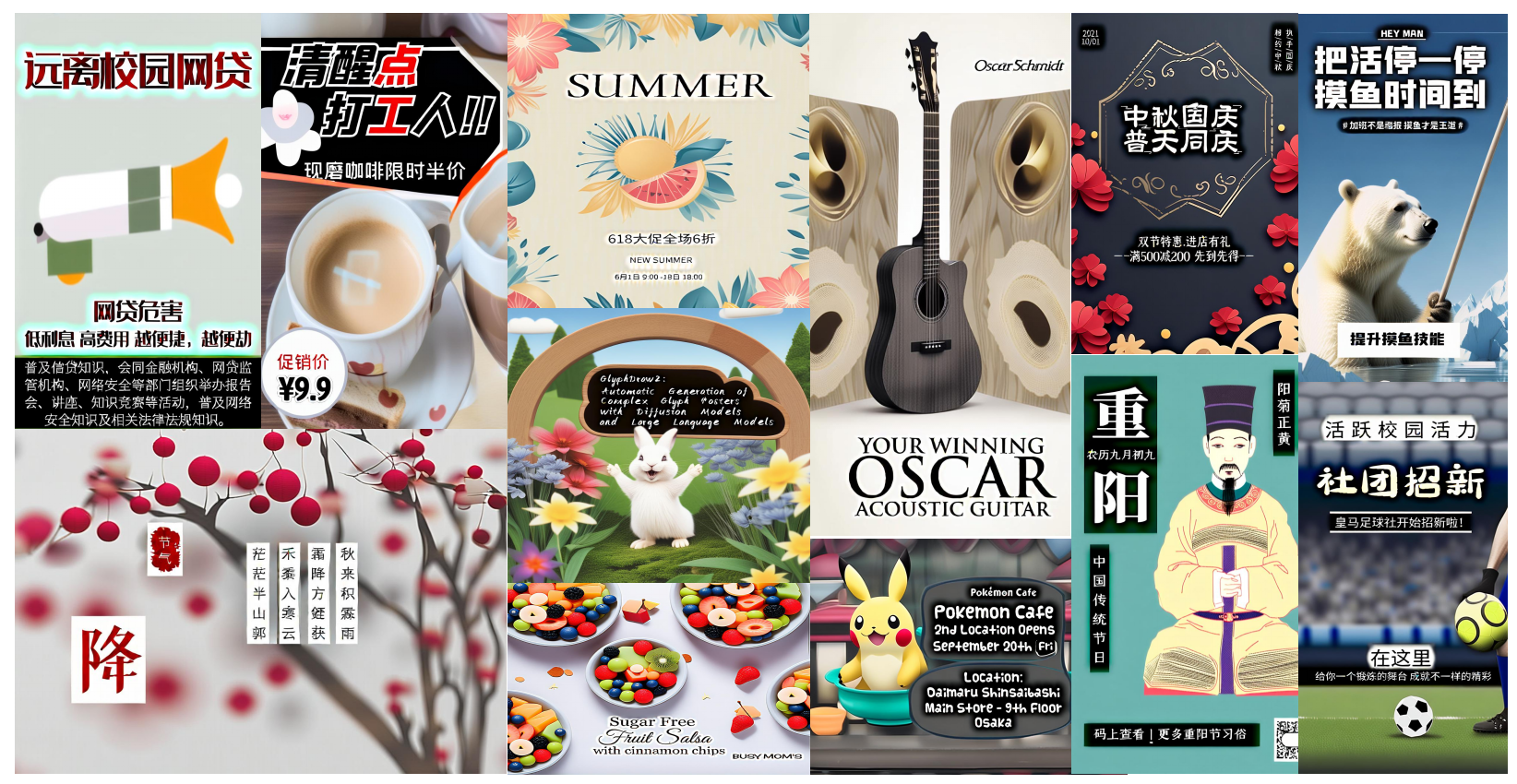}
     \caption{\small{GlyphDraw2 enables seamless and automated generation, eliminating the need for manual box input.}}
\end{figure*}

\begin{abstract}
Posters play a crucial role in marketing and advertising by enhancing visual communication and brand visibility, making significant contributions to industrial design. With the latest advancements in controllable T2I diffusion models, increasing research has focused on rendering text within synthesized images. Despite improvements in text rendering accuracy, the field of automatic poster generation remains underexplored. In this paper, we propose an automatic poster generation framework with text rendering capabilities leveraging LLMs, utilizing a triple-cross attention mechanism based on alignment learning. This framework aims to create precise poster text within a detailed contextual background. Additionally, the framework supports controllable fonts, adjustable image resolution, and the rendering of posters with descriptions and text in both English and Chinese.Furthermore, we introduce a high-resolution font dataset and a poster dataset with resolutions exceeding 1024 pixels. Our approach leverages the SDXL architecture. Extensive experiments validate our method's capability in generating poster images with complex and contextually rich backgrounds.Codes is available at https://github.com/OPPO-Mente-Lab/GlyphDraw2.

\end{abstract}

\section{Introduction}

Posters, as a prominent visual communication medium, have an increasing demand for personalization and customization in various fields of industrial design, whether in advertising, propaganda, marketing or other areas. Although the powerful generative capacity of large-scale T2I diffusion models \cite{nichol2021glide, ramesh2022hierarchical, rombach2022high, saharia2022photorealistic,podell2023sdxlimprovinglatentdiffusion}  enables the creation of images with striking realism and detail, and much research effort has been devoted to addressing the limitations of text rendering in images generated by diffusion models, research on automated poster generation is still relatively limited. The goal of this paper is to endow the diffusion system with the ability to automatically generate posters. We are confronted with three important issues: 1) How can we ensure the accurate generation of the details of paragraph-level small text? 2) How can we ensure the richness of the poster background? 3) How can we eliminate manual input from users and automatically generate posters based on implicit user input?

\begin{figure}[tb]
	\centering
    \includegraphics[width=0.5\textwidth]{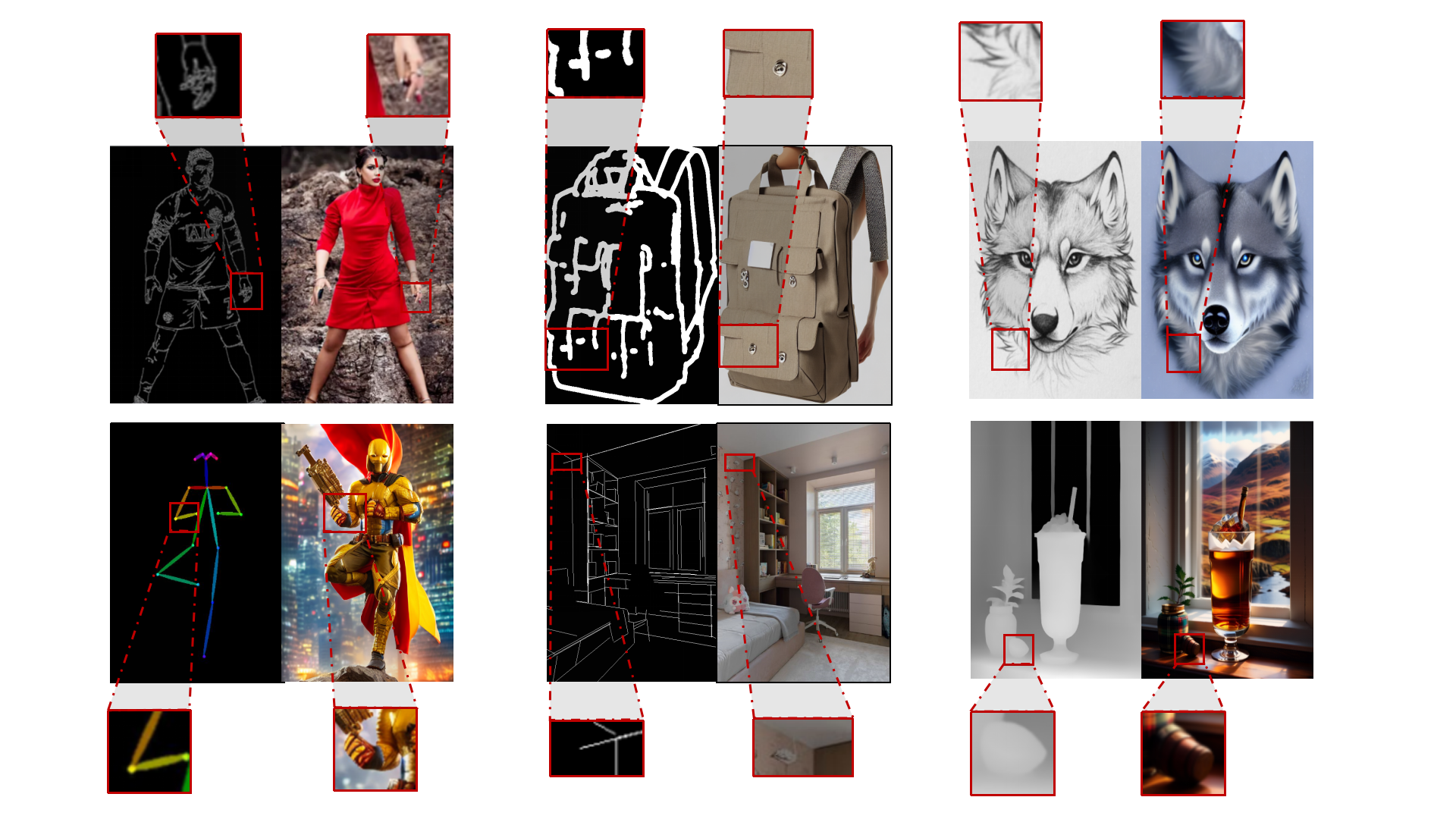}
     \caption{\small{The issues with conventional ControlNet in detail generation.}}
  \label{ControlNet}
\end{figure}

Most recent work on visual text rendering is based on the ControlNet framework\cite{yang2024glyphcontrol,tuo2023anytext,zhao2023udifftext}, which typically uses glyph reference images and specific textual layout details to guide the generation process. However, as shown in Fig.\ref{ControlNet}, the traditional ControlNet has a weak control capability for the minute details, and the paragraph-level small text in posters contains a lot of fine-grained information. In contrast to the global conditional control of traditional ControlNet, control of the font exists only in specific areas and generally accounts for a small proportion of the total image pixels, which is more of a local control. To address these differences, we propose a triple cross-attention method. In addition to the standard cross-attention computation in Unet for interaction between image latent and semantic information, we introduce two additional cross-attention. The interaction Q information for these comes from the image's latent, with the K,V interaction information coming from one feature gained from the font image after glyph encoding and being inserted only into the block corresponding to the SD decoder layer. The goal of this is to learn the font feature detail information and improve the rendering accuracy of small text. The other K,V comes from the features of ControlNet, whose purpose is to adaptively learn the conditional information, that is, the harmony of the font in the overall layout.
Furthermore, in order to ensure the richness of the generated poster background, we introduced an additional alignment target learning. Although multiple control conditions have been introduced, the goal still aligns with the background output of the original prompt semantic condition, so as to keep the model "true to its original aspiration", while ensuring the accuracy of font generation and the richness of the background.
Finally, in order to automatically generate layout conditions and save the cost of manual participation, we construct detailed instruction data ourselves and fine-tune multiple open-source large language models(LLMs) to ensure a seamless user experience during the inference stage.
It is worth mentioning that current LLMs are not yet capable of predicting personalized fonts and colors, but our framework design inherently supports these personalized inputs. Therefore, whether to automate the output depends on the user's decision.

In addition to the above, an essential condition for poster generation is high resolution, with a configurable aspect ratio. Therefore, our framework is based on SDXL adjustments. This brings challenges of difficult data collection amount and quality. Therefore, in addition to collecting open-source data, we specifically completed poster data collection, constructing higher-quality data to support SDXL training. Further, to ensure that the model can simultaneously understand prompts and texts to be rendered in both Chinese and English, we use a PEA\cite{ma2023pea} strategy for multilingual adaptation of the open-source English version of SDXL. The collected font data also includes both Chinese and English.

In summary, our contributions are threefold.
\begin{itemize}
    \item We propose a new model framework for automatically generating poster images. First, we fine-tune Large Language Models (LLMs) to autonomously generate font layouts. We then design a triple cross-attention mechanism to improve the accuracy and controllability of rendered text. Finally, we deploy a semantic alignment module to enhance the richness of the background.
    \item We introduce a high-resolution dataset consisting of fonts with various aspect ratios and poster data. Simultaneously, we design two different types of evaluation benchmarks.
    \item Our final results can achieve certain attribute controls, including font diversity and color control. Both quantitative and qualitative experimental results demonstrate the excellent performance of our proposed architecture in generating posters.
\end{itemize}




\section{Related Work}

\textbf{Controllable Text-to-Image Diffusion Models} Text-to-image (T2I) diffusion models have achieved state-of-the-art results in image generation. While text-based conditioning has improved controllable generation, it doesn't fully meet all user needs. Hence, recent research focuses on adding new types of conditioning to T2I models to address more specific user requirements. One popular method is model-based conditioning, which uses an auxiliary model to encode new conditioning factors, integrating these features into the diffusion model.For example, IP-Adapter\cite{ye2023ip} introduces a decoupled cross-attention mechanism to handle text and image features separately, effectively improving image-based conditioning and influencing subsequent research\cite{ma2023subject, wang2024instantid}. ControlNet\cite{zavadski2023controlnet} is another influential approach, incorporating an additional encoder into the U-Net structure connected via zero convolution. This prevents overfitting and catastrophic forgetting, enabling ControlNet to use specific task inputs as prior conditions for controlled generation. It has been extensively studied for applications like spatial control\cite{jia2024ssmgspatialsemanticmapguided, qin2023unicontrol}, text rendering\cite{yang2024glyphcontrol, zhang2023brushtextsynthesizescene}, and 3D generation\cite{chen2023control3d, yu2023points}.

\textbf{Text Rendering} 
Since GlyphDraw\cite{ma2023glyphdraw} work on Text Rendering last year, numerous excellent follow-up works have emerged. Here, we categorize these works into four groups.
The first group focuses on optimizing Text Rendering accuracy and background coherence. GlyphDraw directly learns by fusing font and text features into a diffusion model, while the subsequent TextDiffuser\cite{chen2024textdiffuser}  adds a Layout Generation module and Character-aware Loss. GlyphControl\cite{yang2024glyphcontrol} introduced ControlNet for Text Rendering, and AnyText\cite{tuo2023anytext} further incorporated auxiliary conditions like text glyph, position, and masked image, as well as a text perceptual loss. Brush Your Text\cite{zhang2023brushtextsynthesizescene}proposes a local attention constraint in the cross-attention layer to address the issue of unreasonable position placement for scene text.
The second group takes the optimization from the perspective of character-aware text encoders. UDiffText\cite{zhao2023udifftext} designs and trains a lightweight character-level text encoder to replace the commonly used CLIP encoder, and Glyph-ByT5 \cite{liu2024glyph} further fine-tunes a character-aware ByT5\cite{xue2022byt5tokenfreefuturepretrained} encoder aligned with glyph features. The motivation behind this group of methods\cite{wang2024instantidzeroshotidentitypreservinggeneration} is derived from personalized generation, the use of dedicated encoders for different categories of conditions tends to significantly improve the final results. DreamText\cite{wang2024highfidelityscenetext} jointly trains text encoders and generators to comprehensively learn and utilize various fonts found in the training dataset. SceneTextGen\cite{zhangli2024layoutagnosticscenetextimage} employs a character-level encoder to extract detailed character-specific features.
The third group primarily considers the text layout, text color, and other high-level image attributes of the generated images. TextDiffuser-2\cite{chen2023textdiffuser} and ARTIST\cite{zhang2024artistimprovinggenerationtextrich} employ LLMs to predict a font's layout. Refining Text-to-Image Generation\cite{lakhanpal2024refiningtexttoimagegenerationaccurate} adopts a text layout generator, and Glyph-byt5 incorporates font type and color control when undertaking Glyph-Alignment Pre-training. CustomText\cite{paliwal2024customtextcustomizedtextualimage} and SceneTextGen similarly consider a variety of text attribute controls. 
The last group involves optimizing the base model from the perspective of data during training. These works\cite{esser2024scalingrectifiedflowtransformers,kolors} generally produce images with strong coherence, but the accuracy of character generation is often relatively low, and the number of characters is highly constrained.

\textbf{LLMs-Generated Text-to-Image Conditions} 
Recent studies \cite{nie2024compositional, zhang2023controllable, gani2023llm} have explored the use of LLMs to generate new comprehensive conditions based on user prompts, such as blob representations, sketches with descriptions, object descriptions, and layout specifications to guide image generation. Especially for layout, LayoutGPT \cite{feng2024layoutgpt} and LayoutPrompter \cite{lin2024layoutprompter} leverage LLMs to generate style sheet language for each object, such as CSS, HTML, XML, ect. Furthermore, TextDiffuser-2, LLM Blueprint \cite{gani2023llm} and Reason Out your Layout \cite{chen2023reason} have explored utilizing LLMs to generate a bounding boxes(bbox) for each object as a new condition. Generating layout bboxes can be achieved through two main approaches: prompt engineering for advanced proprietary models such as GPT-4, and fine-tuning open-source LLMs. Compared to prompt engineering, fine-tuning LLMs is more efficient and facilitates the development of automatic poster generation models.
Based on the above, we fine-tune LLMs on poster layout information to generate bboxes that guide the positioning of textual elements within posters.

\section{Methodology}

\begin{figure}[tb]
	\centering
	\includegraphics[width=0.5\textwidth]{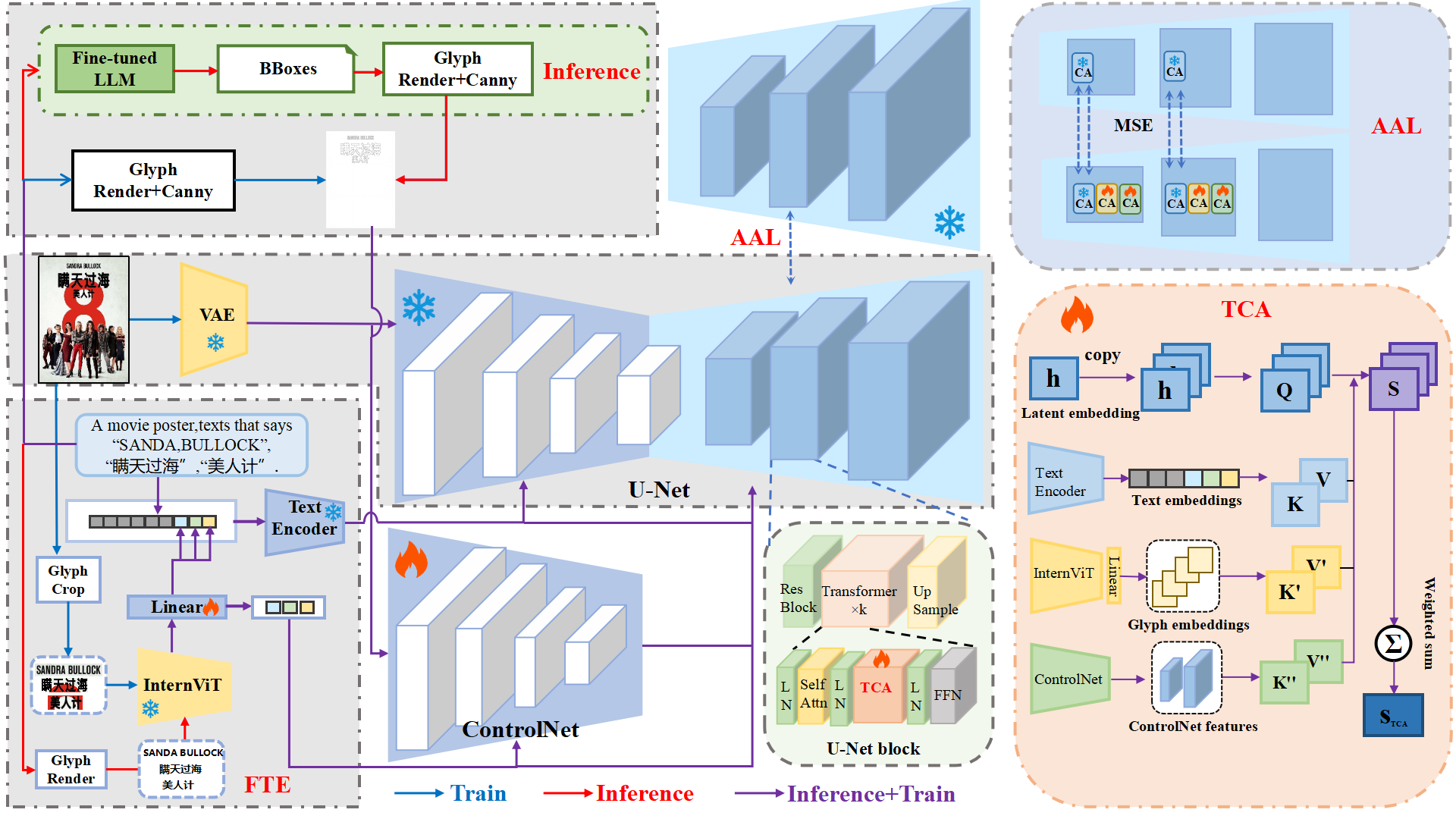}
 \caption{\small{An overview of the proposed GlyphDraw2 method.}}
\label{fig:Framework}
\end{figure}

\subsection{Model Overview}
The entire framework is divided into four parts, as shown in Fig.~\ref{fig:Framework}. The first component, the Fusion Text Encoder (FTE) with glyph embedding, operates in a relatively traditional manner. Its primary objective is to integrate the features of two modalities from the perspective of the text encoder in SD, thereby ensuring a cohesive amalgamation of the two modalities in the generated images. 
The second, and more pivotal, element of our framework is the introduction of Triples of Cross-Attention (TCA). In this stage, we have incorporated two distinct cross-attention layers into the SD decoder section. The first new cross-attention layer facilitates the interaction between glyph features and the hidden variables within the image. This enhancing the accuracy of glyph rendering. Meanwhile, the second new cross-attention layer enables interaction between ControlNet features and the hidden variables in the image. By engaging with ControlNet information, this layer adaptively learns intrinsic data, such as the harmonious layout of the glyph.
In the third part, we have added learning of Auxiliary Alignment Loss (AAL) for semantic consistency, in order to enhance the overall layout and enrich the background information of the poster. 
Finally, in the inference stage, we employed the fine-tuning LLMs strategy to automatically analyze user descriptions and generate corresponding glyphs and coordinate positions of the condition framework. This aims to satisfy automatic poster generation.

\subsection{Fusion Text Encoder}
This approach draws on ideas from earlier works such as Blip-Diffusion \cite{li2024blip}, Subject-Diffusion \cite{ma2023subject}, AnyText, and is also commonly used as a global condition control strategy. Unlike previous methods, we utilized InternViT\cite{chen2023internvl}, a more powerful image encoder specifically trained for character data.
First, the input glyph condition is rendered into a glyph image, then transferred into InternViT to extract corresponding glyph's features. Following the same logic as AnyText, the glyph feature will go through a linear layer for feature alignment when fused with the corresponding position's caption, this ensures the functional modularity of the plug and play, without fine-tuning the text encoder.

\subsection{Triples of Cross-Attention}
In order to ensure the accuracy of glyph generation, we still introduce a ControlNet module here. However, instead of directly adding features in the decoder as before, we additionally introduce a new adaptive cross-attention layer after the original cross-attention layer, as shown in Fig.~\ref{fig:Framework}. 
The output of new cross-attention $S^{'}$ is computed as follows:

\begin{equation}
\begin{aligned}
    &{S^{'}} = {Attention}(Q, K^{'}, V^{'}) = {softmax}\left(\frac{QK^{'T}}{\sqrt{d}}\right)\cdot V^{'},
\end{aligned}
\end{equation}
where $K^{'}={W^{'}_k}^{(j)} \cdot C^{'}$, $V^{'}={W^{'}_v}^{(j)} \cdot C^{'}$, and the $C^{'}$ features come from the corresponding block of ControlNet, ${W^{'}_k}^{(j)}, {W^{'}_v}^{(j)}$ are learnable projection matrices, $j$ represents the block in the U-Net decoder. Due to the asymmetric structure of SDXL's encoder and decoder layers, we have ignored the interaction of the first block in the first two decoders.
The purpose of this approach is that, since the glyph condition only occupies a smaller proportion of the generated image, we need to prevent the ControlNet of the input glyph condition from affecting the richness of the generated image's background. Therefore, we enable adaptive local position learning to ensure glyph condition accuracy while generating images with better layouts and backgrounds.

Moreover, it is worth noting that we have borrowed the approach of InstantID \cite{wang2024instantid}, where the input condition of the ControlNet only contains glyph information, excluding text information.

Furthermore, the accurate generation of paragraphs or larger blocks of text remains a significant challenge. To address this issue, we introduce a second cross-attention layer, 
the output of the second new cross-attention $S^{''}$ is computed as follows:
\begin{equation}
\begin{aligned}
    &{S^{''}} = {Attention}(Q, K^{''}, V^{''}) = {softmax}\left(\frac{QK^{''T}}{\sqrt{d}}\right)\cdot V^{''},
\end{aligned}
\end{equation}
where $K^{''}={W^{''}_k}^{(j)} \cdot C^{''}$, $V^{''}={W^{'}_v}^{(j)} \cdot C^{''}$, and the $C^{''}$ come from the glyph features obtained by InternViT, ${W^{''}_k}^{(j)}, {W^{''}_v}^{(j)}$ are learnable projection matrices,
This idea is inspired by the earlier work of IP-Adapter. It is worth noting that here we also specifically insert this cross-attention layer into the corresponding block of the SD decoder layer only, as modifying the encoder layer would disrupt the features obtained by the ControlNet. Through multiple experiments, we find that the functioning of the ControlNet is highly dependent on its relatively intact encoder structure. Moreover, it is crucial that the ControlNet maintains a duplicate of the SD encoder and uses zero initialization.

In combination with the existing cross-attention layer of each block, the final TCA output is the sum of the three layers as follows:

\begin{equation}
\label{eq:sd}
\begin{aligned}
    &S_{TCA} = \alpha S + \beta S^{'}  + \gamma S^{''},
\end{aligned}
\end{equation}

 where $\alpha,\beta,\gamma$ constants to balance the importance of the three cross-attention layers.
 
\subsection{Auxiliary Align Loss}
Considering the application context for poster generation in our paper, in addition to the accuracy of glyph generation and the harmony of the background, we also need to focus on the richness of the image background itself. Our approach inevitably introduces additional condition injection, including the ControlNet feature addition as well as the TCA strategy which results in an increased number of decoder components. The fundamental purpose of these conditions is to ensure the controllability of the generated image. However, many articles have shown that controllability is often accompanied by a sacrifice in editability or text consistency. Therefore, we introduce AAL in our approach. The alignment model employs SDXL as its backbone, similar to how ControlNet utilizes a duplicated SD encoder. However, in our method, we duplicate the SD decoder and apply AAL between the cross-attention outputs in each block of the duplicated decoder and those in the original cross-attention layer of the TCA. The primary objective of this approach is to minimize the impact of the added modules for learning glyphs on the overall layout and image quality.
Therefore, our AAL for semantic consistency $\mathcal{L}'$ can be formulated as follows:

\begin{equation}
\mathcal{L}' = \lVert  {softmax}\left(\frac{QK^{T}}{\sqrt{d}}\right)\cdot V -  {softmax}\left(\frac{QK_c^{T}}{\sqrt{d}}\right)\cdot V_c \rVert,
\end{equation}

where $K_c,V_c$ refers to the CA output in each block of the duplicated U-Net decoder. Our final loss can be formulated as follows with an important hyperparameter $\lambda$:

\begin{equation}
\mathcal{L} = \mathbb{E}_{\mathcal{E}(x_0),C,\epsilon \sim \mathcal{N}(0,1),t} \Big[ \lVert \epsilon - \epsilon_\theta (z_t, t, C)\rVert_2^2 \Big] + \lambda \mathcal{L}'.
\end{equation}

\subsection{Inference with Fine-tuned LLMs}
To ensure automatic poster generation, the last problem that urgently needs to be solved is the elimination of manual intervention, i.e., the process of predefined image layout. We rely completely on user's caption description and introduce LLMs to solve this problem. Also, for the convenience of invocation, we have constructed our own instruction data and fine-tuned the open-source language model.

\section{Experiments}

\subsection{Implementation Details}

The model we intend to train comprises two main components. The first component is a controllable T2I poster model, with the backbone of our framework being based on SDXL. To adapt the multilingual understanding capacity of the SDXL encoder and maintain linguistic coherence between the prompt's description of the poster background and the generated text, we have incorporated the PEA-Diffusion strategy \cite{ma2023pea} into the backbone architecture. The second component is a layout generation model based on LLMs. For more training details, please refer to the Appendix.

\subsection{Evaluation}
The evaluation set can be divided into five parts, which are used to assess the performance of the model.

\textbf{AnyText-Benchmark}\cite{tuo2023anytext} contains one thousand English images and Chinese images from LAION \cite{schuhmann2021laion} and Wukong \cite{gu2022wukong} respectively. 

\textbf{ICDAR13}\cite{6628859} serves as the benchmark for assessing the detection of near-horizontal text, and it consists of 233 images for testing purposes.Using this evaluation set also provides a better comparison with the UDiffText.

\textbf{MARIO-Eval}\cite{chen2023textdiffuser} serves as a comprehensive tool for evaluating text rendering quality collected from the subset of MARIO-10M test set and other sources. Using this evaluation set also provides a better comparison with the TextDiffuser.

It is worth noting that in the AnyText-Benchmark, ICDAR13, and MARIO-Eval, the majority of English evaluation sets contain only a single English word per bbox. This often results in a lack of precision when evaluating complete English sentences. Consequently, there is a pressing need to construct more complex and comprehensive evaluation sets to improve accuracy.

\textbf{Complex-Benchmark} consists of 200 prompts which include bilingual Chinese and English. In the Chinese prompts, the characters to be rendered are randomly combined with intricate strokes and structures, while the English prompts feature longer words with consecutive repetitions of letters. 
For more details, please refer to the Appendix.

\textbf{Poster-Benchmark} includes 240 prompts that describe the generation of posters.
For more details, please refer to the Appendix.
Its purpose is to evaluate the layout accuracy, robustness, and overall aesthetic quality of automatic poster generation.

\textbf{Evaluation Metrics.} For these evaluation sets, we utilized four evaluation metrics to assess the accuracy and quality of poster generation: (1)\textbf{Accuracy (Acc)} calculates the proportion of correctly generated characters in the rendered text compared to the total number of characters that need to be rendered.(2)\textbf{Normalized Edit Distance (NED)},the calculation method remains consistent with AnyText.
(3) \textbf{ClipScore} measures how well the generated image aligns with the textual prompt or description provided. 
(4)\textbf{HPSv2} \cite{wu2023humanpreferencescorev2} whether the generated images align with human preferences and serves as an indicator to assess the preferences quality of the images.

In our comparison, we evaluated various methods, mainly including three categories. The first category is the recently open-sourced large-scale text generation models with font rendering support from the industry, including StableDiffusion3 (SD3)\cite{esser2024scalingrectifiedflowtransformers}, Kolors\cite{kolors}, and FLUX.1 series made by Black Forest Labs. Among them, SD3 and FLUX.1 only support English. Besides, NED calculations usually rely on anchoring based on text box positioning, so the NED calculations for this category are ignored.
The second category is open-source font rendering text generation methods, which include TextDiffuser series, AnyText, UDiffText, and Glyph-ByT5.
The third category is the comparative experiments based on the basic ControlNet, which include two different conditional inputs. One is the early version of related methods that directly use rendered fixed-font as the condition\cite{ma2023glyphdraw,yang2024glyphcontrol,tuo2023anytext}, and the other is based on the canny aspect of the original font of the training images as the condition.

\subsection{Experimental Results}
In the following section, we provide a comprehensive analysis of both quantitative and qualitative results, comparing our method with state-of-the-art approaches in the fields of text rendering and poster generation. 

We record all the comparative experiments in Table\ref{table1}. GlyphDraw1.1 implies that the conditional input of ControlNet and the input of InternViT are rendered images of fixed fonts. GlyphDraw2 indicates that the conditional input of ControlNet is the canny image of the corresponding real font in the picture, and the input of InternViT is the actual font of the corresponding picture, that is, the overall framework diagram in Fig.\ref{fig:Framework}.
Additionally, the calculation of accuracy in AnyText-Benchmark uses the PWAcc indicator that calculates the accuracy of the words generated at a specific position, while the Acc indicator is used in other evaluation sets.

\begin{table*}[ht]
\centering
\caption{Evaluation Results on five benchmarks.}
\label{table1}
\resizebox{0.9\linewidth}{!}{
\begin{tabular}{cccccccccc}
\hline
\multirow{2}{*}{\begin{tabular}[c]{@{}c@{}}Evaluation\\ Benchmark\end{tabular}} & \multirow{2}{*}{Model}        & \multicolumn{4}{c}{Chinese}          & \multicolumn{4}{c}{English}          \\ \cline{3-10} 
   &         & Acc  & NED  & ClipScore       & HPSv2& Acc  & NED  & ClipScore       & HPSv2\\ \hline
\multirow{11}{*}{\begin{tabular}[c]{@{}c@{}}AnyText-\\ Benchmark\end{tabular}}  & SD3     & -    & -    & -    & -    & 0.3261          & -    & 0.4517          & 0.2215          \\
   & Kolors  & 0.0665          & -    & \textbf{0.4011} & \textbf{0.2654} & 0.0243          & -    & 0.4854          & 0.2512          \\
   & FLUX.1-schnell     & -    & -    & -    & -    & 0.3884          & -    & \textbf{0.4914} & \textbf{0.2541} \\ \cline{2-10} 
   & ControlNet         & 0.7598          & 0.8254          & 0.3749          & 0.2347          & 0.7098          & 0.8467          & 0.4558          & 0.2245          \\
   & ControlNet w/ canny& 0.7804          & 0.8365          & 0.3752          & 0.2384          & 0.7954          & 0.8745          & 0.4599          & 0.2287          \\
   & TextDiffuser$\dagger$       & 0.0605          & 0.1262          & -    & -    & 0.5921          & 0.7951          & -    & -    \\
   & AnyText-v1.1       & 0.7661          & 0.8423          & 0.3968          & 0.2272          & 0.7108          & 0.8564          & 0.4721          & 0.2121          \\
   & UDiffText          & -    & -    & -    & -    & 0.6435          & 0.8284          & 0.4645          & 0.2214          \\
   & Glyph-ByT5         & 0.7227          & 0.7799          & 0.4005          & 0.2601          & 0.7307          & 0.8353          & 0.4802          & 0.2511          \\
   & \textbf{GlyphDraw1.1 w/o LLMs} & 0.7892          & 0.8476          & 0.3921          & 0.2555          & 0.7369          & 0.8921          & 0.4616          & 0.2350          \\
   & \textbf{GlyphDraw2 w/o LLMs}& \textbf{0.8266} & \textbf{0.8543} & 0.3986          & 0.2589          & \textbf{0.8627} & \textbf{0.9278} & 0.4796          & 0.2451          \\ \hline
\multirow{2}{*}{ICDAR13} & UDiffText          & -    & -    & -    & -    & 0.5840          & 0.7221          & 0.4521          & 0.2101          \\
   & \textbf{GlyphDraw2}& -    & -    & -    & -    & 0.6901          & 0.7629          & 0.4657          & 0.2345          \\ \hline
\multirow{2}{*}{MARIO-Eval}         & TextDiffuser$\dagger\dagger$       & -    & -    & -    & -    & 0.5609          & -    & -    & -    \\
   & \textbf{GlyphDraw2}& -    & -    & -    & -    & 0.7672          & 0.9330          & 0.4765          & 0.2464          \\ \hline
\multirow{14}{*}{\begin{tabular}[c]{@{}c@{}}Complex-\\ Benchmark\end{tabular}}  & SD3     & -    & -    & -    & -    & 0.2515          & -    & \textbf{0.4391} & 0.2492          \\
   & Kolors  & 0.0198          & -    & \textbf{0.3878} & \textbf{0.2546} & 0.0033          & -    & 0.4254          & \textbf{0.2546} \\
   & FLUX.1-schnell     & -    & -    & -    & -    & 0.2969          & -    & 0.4298          & 0.2544          \\ \cline{2-10} 
   & ControlNet         & 0.6943          & 0.8745          & 0.3589          & 0.2364          & 0.2254          & 0.4025          & 0.4214          & 0.2385          \\
   & ControlNet w/ canny& 0.7546          & 0.8812          & 0.3512          & 0.2386          & 0.4215          & 0.4532          & 0.4311          & 0.2298          \\
   & AnyText-v1.1       & 0.5749          & 0.8560          & 0.3633          & 0.2434          & 0.0342          & 0.3755          & 0.4104          & 0.2312          \\
   & Glyph-ByT5         & 0.7895          & 0.8263          & 0.3711          & 0.2455          & 0.4834          & 0.7034          & 0.4256          & 0.2412          \\
   & \textbf{GlyphDraw1.1 w/o LLMs} & 0.7176          & 0.8991          & 0.3600          & 0.2422          & 0.2791          & 0.4332          & 0.4160          & 0.2395          \\
   & \textbf{GlyphDraw2 w/o LLMs}& \textbf{0.9051} & \textbf{0.9037} & 0.3702          & 0.2411          & \textbf{0.5574} & \textbf{0.4928} & 0.4211          & 0.2414          \\ \cline{2-10} 
   & LLMs+ControlNet     & 0.5812          & 0.8012          & 0.3687          & 0.2365          & 0.1856          & 0.5841          & 0.4215          & 0.2356          \\
   & TextDiffuser-2     & -    & -    & -    & -    & 0.0999          & 0.4428          & 0.3985          & 0.2285          \\
   & LLMs+AnyText-v1.1   & 0.4850          & 0.7888          & 0.3697          & \textbf{0.2534} & 0.0455          & 0.4680          & 0.4038          & 0.2380          \\
   & \textbf{GlyphDraw1.1}         & 0.6215          & 0.8479          & \textbf{0.3756} & 0.2427          & 0.2264          & 0.6273          & \textbf{0.4362} & 0.2415          \\
   & \textbf{GlyphDraw2}& \textbf{0.6691} & \textbf{0.7975} & 0.3754          & 0.2498          & \textbf{0.4158} & \textbf{0.6294} & 0.4312          & 0.2488          \\ \hline
\multirow{10}{*}{\begin{tabular}[c]{@{}c@{}}Poster-\\ Benchmark\end{tabular}}   & SD3     & -    & -    & -    & -    & 0.2310          & -    & 0.4128          & 0.2337          \\
   & Kolors  & 0.0426          & -    & \textbf{0.4110} & \textbf{0.2510} & 0.0020          & -    & 0.4120          & 0.2421          \\
   & FLUX.1-schnell     & -    & -    & -    & -    & 0.3744          & -    & \textbf{0.4215} & \textbf{0.2541} \\ \cline{2-10} 
   & ControlNet         & 0.7878          & 0.8453          & 0.3844          & 0.2298          & 0.3421          & 0.7514          & 0.3902          & 0.2125          \\
   & ControlNet w/ canny& 0.7911          & 0.8541          & 0.3801          & 0.2225          & 0.5012          & 0.8014          & 0.3955          & 0.2106          \\
   & TextDiffuser-2     & -    & -    & -    & -    & 0.1046          & 0.3623          & 0.3914          & 0.2110          \\
   & LLMs+AnyText-v1.1   & 0.7421          & 0.8894          & 0.3956          & 0.2362          & 0.2604          & 0.7120          & 0.4093          & 0.2289          \\
   & Glyph-ByT5         & 0.8248          & 0.9040          & 0.4012          & 0.2366          & 0.7341          & 0.8411          & 0.4101          & 0.2354          \\
   & \textbf{GlyphDraw1.1}         & 0.8215          & 0.9590          & 0.3908          & 0.2378          & 0.3999          & 0.7667          & 0.3984          & 0.2297          \\
   & \textbf{GlyphDraw2}& \textbf{0.8263} & \textbf{0.9585} & 0.3987          & 0.2314          & \textbf{0.7590} & \textbf{0.8759} & 0.4114          & 0.2301          \\ \hline
\end{tabular}}
\end{table*}

\textbf{Comparison results of AnyText-Benchmark.} To ensure fair evaluation, all methods employed the DDIM sampler with a sampling step of 50, CFG scale of 9, and a fixed random seed of 100. 
Each prompt generated a single image with identical positive and negative cues. From the results, It is evident that our model achieves significantly higher accuracy in rendering both Chinese and English text compared to AnyText. 
The ClipScore metric is slightly lower than that of GlyphDraw1.1 and closely similar to GlyphDraw2.
Since the weights of TextDiffuser are currently not downloadable, the results of TextDiffuser$\dagger$ come from the AnyText's results, and ClipScore and HPSv2 can't be calculated. UDiffText does not support Chinese, and the open-source weights only support editing, so the metrics tested here are calculated by directly editing the bbox content in AnyText-Benchmark. At the same time, it only supports the generation of limited characters within 12, and does not support the generation of long characters.
Finally, it is worth mentioning that, the ClipScore and HPSv2 metrics are lower compared to Glyph-ByT5, indicating that Glyph-ByT5 indeed has certain advantages in terms of image-text consistency and human preference metrics. However, during our actual subjective testing, we found that the font generated by Glyph-ByT5 sometimes did not generate within the given bbox, indicating a certain degree of uncontrollability.

\textbf{Comparison results of ICDAR13.} 
UDiffText took some restrictions in the testing on the ICDAR13 evaluation set. For instance, the authors chose to edit only a hundred words for evaluation and ignored the case of letters during evaluation. Moreover, the Acc metric they used was character-level, not word-level. We lifted these restrictions and reran the ICDAR13 evaluation set with UDiffText, providing a result comparison. Our results have obvious advantages in four metrics.

\textbf{Comparison results of MARIO-Eval.} 
Similarly here, the result represented by TextDiffuser$\dagger\dagger$ comes from the TextDiffuser itself. Since we can't get the open-source model, we only compared the Acc metrics. Our result has a significant advantage.

\textbf{Comparison results of Complex-Benchmark.} 
Complex-Benchmark does not provide fixed font bbox, so in addition to the comparison experiments of the three large T2I models, we have conducted two types of comparison experiments. One type is based on the character count and size, which randomly assign bbox. The main purpose of this type is to test the upper limit of the accuracy of complex font generation by the model without bbox restrictions. The other type is using the fine-tuned LLMs to predict the rendered characters and their corresponding bbox, testing the complex font generation ability in real-world scenarios. This approach allows for a more in-depth evaluation and comparison of the automatic text generation functionality.

Firstly, in the comparison of the three end-to-end T2I large models, although Kolors already supports Chinese font rendering, after testing, it was found that the ability to generate complex characters is relatively weak, and the Acc is only around 0.02. In the English evaluation set, FLUX.1 with 12 billion parameters has a significant advantage.
Secondly, in experiments with randomly given bbox, GlyphDraw2 still has significant advantages in both Acc and NED metrics in both Chinese and English evaluation sets. In the English evaluation set, the text rendering accuracy of AnyText is quite low. Although the accuracy of GlyphDraw2 is not high, it has far exceeded AnyText. However, the ClipScore metric of GlyphDraw2 is lower than the result of ControlNet. From our investigation, we found that the quality of our English data is relatively low, which might be one of the reasons.
In the Chinese evaluation set, the ClipScore has a slight advantage in comparison with other methods except for Glyph-ByT5.
Finally, as for the results of automatically generating bbox as the condition, since TextDiffuser-2 also automatically predicts the bbox, it is also compared here. TextDiffuser-2 does not support Chinese, and the test results of the English evaluation set metrics are low. After analysis, it was found that TextDiffuser-2's language model predictions for the text and bbox to be rendered have a significant issue of incorrect and missing characters. At the same time, in order to compare with AnyText, we used the bbox generated by our fine-tuned LLMs as the conditional input for AnyText. The conclusion is similar to the above, and our scheme has a significant advantage in accuracy metrics.

\textbf{Comparison results of Poster-Benchmark.} 

The performance of the three end-to-end T2I large models is similar to that of Complex-Benchmark. In addition to having strong text rendering capabilities, FLUX.1 also has significant advantages in image-text consistency and human preference metrics. I believe this is closely related to its 12 billion parameters and the MMDiT\cite{esser2024scalingrectifiedflowtransformers} architecture.
Additionally, it needs to be noted that Glyph-ByT5 has already surpassed GlyphDraw1.1 in terms of metrics, and the Acc metric in the Chinese evaluation set is close to GlyphDraw2. This indicates that there are indeed significant advantages and potential in using the fine-tuned ByT5 as the character encoder.

One last point to note is that there is a significant gap between GlyphDraw1.1 and GlyphDraw2 on the English evaluation sets of Complex-Benchmark and Poster-Benchmark. After offline experimental analysis, it was found that due to the strong diversity of the English fonts we constructed, when we fixed the font group as the condition for the ControlNet, there was a significant gap with the real font, which resulted in a slower fitting process during training. At the same time, it also led to English results significantly lower than the results where the canny font was used as the condition directly.

\begin{figure}[tb]
	\centering
    \includegraphics[width=0.5\textwidth]{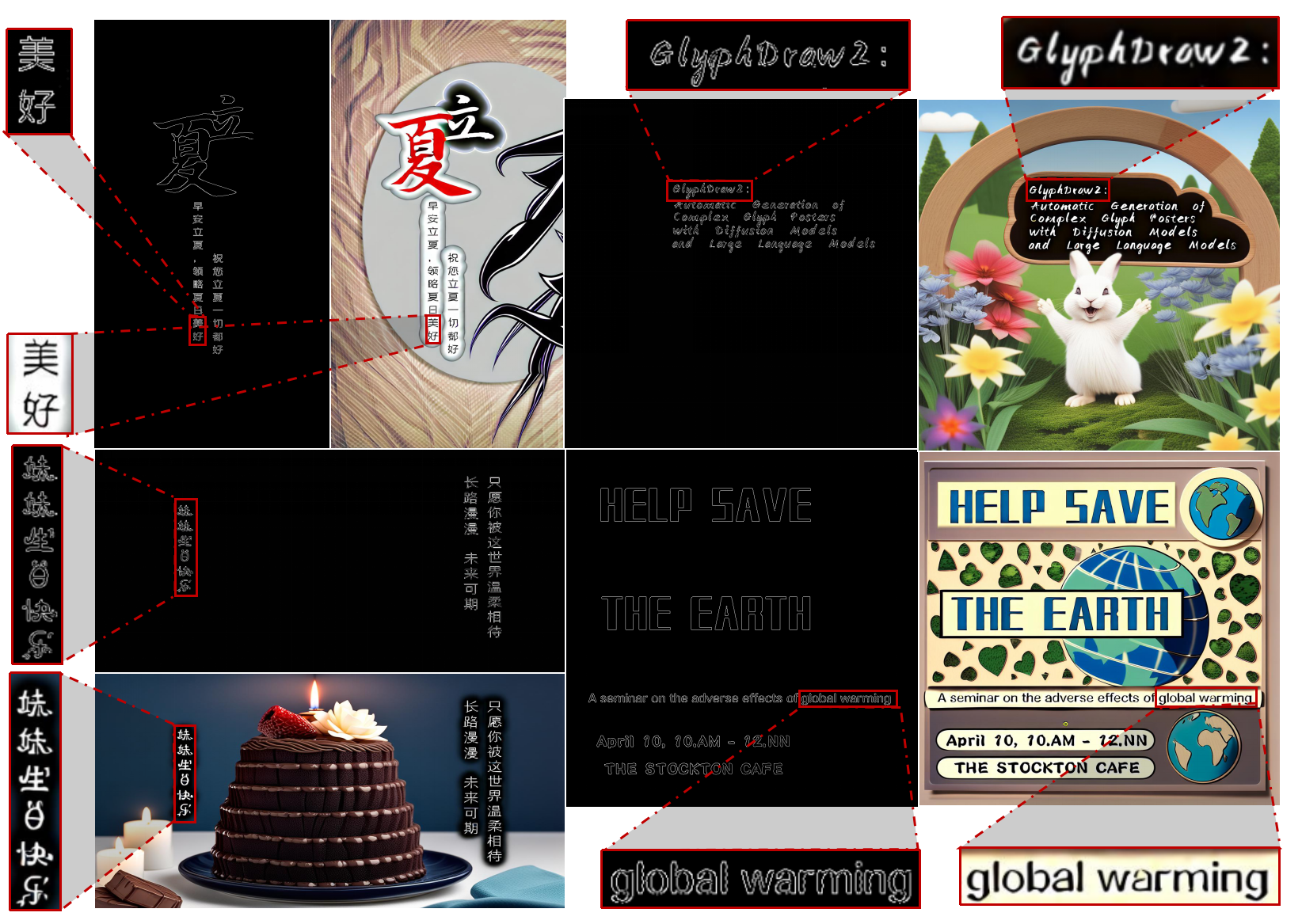}
     \caption{\small{The detail generation with GlyphDraw2.}}
  \label{detail_generation}
\end{figure}

\textbf{Qualitative results.} 
Fig.\ref{detail_generation} shows some magnified detail generation. Compared with the traditional ControlNet results in Fig.\ref{ControlNet}, GlyphDraw2 indeed has certain advantages in detail generation.

\textbf{LLMs layout prediction experiment.} 
We randomly tested 1000 prompts, using the correctness of the predicted format as the basis for calculating accuracy. Although a correctly predicted format does not necessarily mean the real rendering position is correct, this kind of error is relatively minor. Here we select three models for comparison, namely Qwen1.5~\cite{qwen}, Baichuan2~\cite{yang2023baichuan2openlargescale}, and Llama2~\cite{touvron2023llama2openfoundation}. Among them, we experiment three model sizes for Qwen1.5, while the other two models were tested with two model sizes each. For more experimental details, please refer to the Appendix.

\subsection{Ablation Studies}

\begin{table}[ht!]
\centering 
\caption{Ablation Results on Poster-Benchmark in Chinese.}
\label{table2}
\resizebox{1\linewidth}{!}{
\begin{tabular}{cccccccccc}
\hline
\multicolumn{6}{c}{Model}     & \multicolumn{4}{c}{Chinese}   \\ \hline
\begin{tabular}[c]{@{}c@{}}w/ \\CAG\end{tabular}  & \begin{tabular}[c]{@{}c@{}}w/ \\CAC\end{tabular}         & \begin{tabular}[c]{@{}c@{}}w/ \\TCA\end{tabular}         & \begin{tabular}[c]{@{}c@{}}w/ \\AAL\end{tabular}   & \begin{tabular}[c]{@{}c@{}}w/ \\FTE\end{tabular} & \begin{tabular}[c]{@{}c@{}}w/ \\CC\end{tabular} & Acc   & NED   & ClipScore       & HPSv2 \\ \hline
         &      &      &&       &      & 0.7782& 0.9396& 0.4098& 0.2464\\
 \Checkmark       &      &      &&       &      & 0.8014& 0.9548& 0.3968& 0.2365\\
         & \Checkmark    &      &&       &      & 0.7845& 0.9354& 0.4104& \textbf{0.2488} \\
 \Checkmark       & \Checkmark    & \Checkmark    &&       &      & 0.8154& \textbf{0.9588} & 0.4099& 0.2484\\
         &      &      & \Checkmark        &       &      & 0.7689& 0.9245& \textbf{0.4121} & 0.2455\\
 \Checkmark       & \Checkmark    & \Checkmark    & \Checkmark        &       &      & 0.8122& 0.9451& 0.4108& 0.2444\\
         &      &      && \Checkmark     &      & 0.7841& 0.9314& 0.4067& 0.2476\\
 \Checkmark       & \Checkmark    & \Checkmark    & \Checkmark        & \Checkmark     &      & \textbf{0.8161} & 0.9511& 0.4099& 0.2401\\
         &      &      &&       & \Checkmark    & 0.7854& 0.9254& 0.3987& 0.2387\\ \hline
\begin{tabular}[c]{@{}c@{}}w/ \\ByT5\end{tabular} & \begin{tabular}[c]{@{}c@{}}w/ \\CB\end{tabular} & \begin{tabular}[c]{@{}c@{}}w/ \\FT\\ CB\end{tabular} & \begin{tabular}[c]{@{}c@{}}w/ \\PP-\\OCR\end{tabular} & \multicolumn{2}{c}{w/ InternViT} & Acc   & NED   & ClipScore       & HPSv2 \\ \hline
\Checkmark       &      &      && \multicolumn{2}{c}{}   & 0.7951& 0.9125& 0.3996& 0.2361\\
         & \Checkmark    &      && \multicolumn{2}{c}{}   & 0.7981& 0.9025& 0.4012& 0.2341\\
         &      & \Checkmark    && \multicolumn{2}{c}{}   & 0.8017& 0.9147& 0.4004& 0.2334\\
         &      &      & \Checkmark        & \multicolumn{2}{c}{}   & 0.8014& 0.9245& 0.3996& 0.2302\\ \cline{7-10} 
        &      &      && \multicolumn{2}{c}{\textbf{\Checkmark}}   & \textbf{0.8263} & \textbf{0.9585} & \textbf{0.3987} & \textbf{0.2314} \\ \hline
\end{tabular}}
\end{table}

Our ablation experiments are mainly divided into two parts. The first part adopts a comparison strategy of adding modules to the ControlNet base model and mainly includes four main aspects: 1) the impact of TCA and its specific modules; 2) the impact of AAL; 3) the impact of text encoder fusion; and 4) the impact of ControlNet's condition input.
The second part of the experiment involves ablating and comparing the glyph encoder structure of InternViT from Fig.~\ref{fig:Framework}. Since the output of the glyph encoder greatly affects the entire framework and serves as a part of the input for the FTE module, ControlNet's conditional input, and the input for the CAG module within the TCA module, we conducted many experiments to demonstrate that the encoding capability of the glyph encoder is positively correlated with the overall model performance, further substantiating the effectiveness of the overall framework. Four main experiments were conducted here: 1) Directly encoding the text to be rendered using ByT5, as ByT5 eliminates the need for SentencePiece vocabulary by directly inputting UTF-8 bytes into the model without any text preprocessing; 2) Directly encoding the text to be rendered using ChineseBERT(CB)\cite{sun2021chinesebertchinesepretrainingenhanced}, which combines glyph embedding, pinyin embedding, and character embedding information; 3) Building on experiment 2, we borrowed ideas from papers such as UDiffText and Glyph-ByT5 to fine-tune the CB using the CLIP model framework, with the image side employing PP-OCR; 4) Rendering the text into image information first, then encoding it using PP-OCR.

\textbf{Effectiveness of TCA}: TCA adds two CA layers, and we experimentally verify each added CA layer separately. Here, w/ CAG refers to the ablation experiment where font features are used as K, V for CA interaction. Since the CA layer is added to improve the font accuracy, as shown in Table~\ref{table2}, the addition of this layer improves ACC and NED but causes a certain decrease in ClipScore and aesthetic metrics, indicating that the addition of CAG improves font rendering accuracy while sacrificing some text semantic alignment capabilities. CAC represents the experimental process of adaptive feature interaction in CA, which stems from the ControlNet encoder's features. Apart from the NED metric, other metrics show a slight improvement, indicating that adaptive feature interaction can indeed enhance both the accuracy of font rendering and the text semantic alignment ability, as well as aesthetic metrics. w/ TCA refers to the experiment carried out with the entire TCA module, where ACC, NED, and other metrics show a certain improvement. This further illustrates that the TCA module plays a positive role in improving the accuracy of font rendering and the aesthetic score of images.

\textbf{Effectiveness of AAL}: w/ AAL represents the ablation experiment for the align semantic strategy. As can be seen from Table~\ref{table2}, this strategy does improve the semantic alignment ability and the image quality to some extent, albeit sacrificing some font rendering accuracy. However, the overall impact remains positive. Further, as can be seen from the 6th row of the first half of Table~\ref{table2}, adding the TCA strategy and the AAL strategy simultaneously generates a significant improvement in metrics compared to adding the AAL strategy alone. However, compared to just adding the TCA strategy, there is no significant difference in Acc, NED, and HPSv2 metrics, and there is a certain increase in Clipscore, indicating that the superposition of these two strategies has some effect on improving image-text consistency.

\textbf{Effectiveness of FTE}: The primary purpose of FTE is to enhance the harmony between the font and background. As can be observed from the ablation study in Table~\ref{table2}, all metrics are affected to some degree. FTE incorporates font feature information, which enhances text rendering accuracy. However, fusing image modalities may weaken the alignment of text semantics, resulting in a slight decline in ClipScore. Finally, enhancing image compatibility positively affects the preference score. Similarly, in the eighth row of the upper half of Table~\ref{table2}, the TCA and AAL strategies are cumulatively added, and the overall effect is positive.

\textbf{Effectiveness of ControlNet's condition input}: CC signifies that ControlNet's condition input includes only the font image features, reducing the impact of the descriptive caption on font rendering. This somewhat improves the font accuracy.

\textbf{The Role of the Glyph Encoder}: In this part, five experiments were conducted, comparing different encoders within the overall framework. There are two core conclusions. Firstly, the effect of directly using a text encoder is not as good as employing a visual encoder, even when fine-tuning the text encoder within a comparative learning framework. Secondly, when the visual encoder has a more significant capacity and has been extensively trained on images related to text, it performs better than the conventional OCR encoder.

\section{Conclusion and Limitation}
So far, the profound cost and limited availability of manual labeling have presented significant challenges to the practical deployment of glyph generation models. In this study, we first collected high-resolution images containing Chinese and English glyphs and subsequently constructed an automatic screening process to build a large-scale dataset. Subsequently, we establish a comprehensive framework that merges text and glyph semantics, leveraging various tiers of information to optimize text rendering accuracy and richness of the background.
Empirical analysis from our experiments demonstrates that our methodology surpasses existing models on various evaluation sets, suggesting potential to serve as a foundation for enhancing automatic poster generation capabilities.

\paragraph{Limitation}
Although our method can generate automatic posters of free resolution, there are still some issues at present. Firstly, for the glyph bboxes predicted by LLMs, the prediction accuracy is meager for complex scenarios, such as when a user inputs a paragraph of text without quotation marks as a bbox prompt. Secondly, balancing the richness of background generation and the accuracy of text rendering is still relatively difficult. In our current approach, we prioritize glyph accuracy; thus, the visual appeal of the background may be weaker. Additionally, the generation accuracy for tiny glyphs or paragraph texts still needs improvement. In the future, we may explore some solutions on the text encoder side to address these issues.

\clearpage

\bibliography{aaai24}

Appendix~\ref{data} explains the organization and preparation of the training data. 

Appendix~\ref{Implementation} details the training procedures for each stage. 

Appendix~\ref{Self-Constructed} describes the two evaluation benchmarks in detail. 

Appendix~\ref{GlyphDraw1.1} presents subjective results and an ablation study using GlyphDraw1.1. 

Appendix~\ref{LLMs} outlines the construction of instruction data for LLM fine-tuning and experiments with various models. 

Appendix~\ref{Style} showcases the controllable generation results of the model in different fonts. 

Appendix~\ref{subjective} includes additional comparative experiments on subjective results. 

Appendix~\ref{related} summarizes the font rendering work comprehensively.

Appendix~\ref{Limitations_a} discusses the limitations of GlyphDraw2 and future research prospects.

\section{Dataset}
\label{data}

\subsection{Motivation}
To endow the diffusion model with the ability to produce effective poster images, the construction of a comprehensive dataset with the following characteristics is necessary: diverse glyph distributions, aesthetically pleasing layouts and compositions, and visually appealing backgrounds.
In addition, our specific objective focuses on achieving versatility through bilingual poster generation.
However, existing datasets mainly emphasize text-image datasets specifically tailored for monolingual text rendering, such as LAION-Glyph \cite{yang2024glyphcontrol} and MARIO-10M \cite{chen2024textdiffuser} for English generation, which also demonstrate slight limitations in terms of text layout.
AnyWord-3M \cite{tuo2023anytext} is a bilingual dataset predominantly sourced from e-commerce and advertising contexts. While it is well-suited for text rendering training, it lacks sufficient text layout and background appeal for poster generation tasks, making it less than ideal as a standalone training dataset for poster generation.


\subsection{Data Collection and Processing}

We create two large-scale high-resolution image datasets to improve the precision of the generated text and enhance the overall visual quality in the poster generation task. These datasets boast resolutions exceeding 1024x1024, facilitating the production of more accurate and aesthetically pleasing posters. The first dataset, known as the general dataset, is intended to train the model's text rendering capabilities. The second dataset is specifically designed for poster generation and includes mainly Chinese glyphs in the text of poster images, with about 10\% of the data consisting of English words. To obtain high-quality images embedded with visual text for poster generation tasks, we employ a data preprocessing procedure to filter data and extract text and location information.


The initial step entails processing the general dataset. Specifically, high-resolution images are first selected, as our chosen base model is SDXL. Following this, PP-OCR is employed to precisely locate and recognize text elements within the images, encompassing both English and Chinese characters. In order to mitigate potential noise in the dataset, we employ sophisticated filtering strategies specifically designed for the identified text bboxes. Additionally, we leverage the BLIP-2 \cite{li2023blip} to generate captions for the collected images. The extracted text is enclosed within quotation marks and seamlessly integrated into the image captions. 

For the poster dataset, supplementary resolution-based filtering rules are introduced to carefully select landscape and portrait-oriented posters. Furthermore, aesthetic scoring is utilized to identify visually captivating images of higher quality. In addition, to enhance the dataset's overall quality, improvements were made to the image processing techniques. Due to the potential limitations of PP-OCR in accurately locating and extracting all text and its corresponding positions in images, the unrecognizable portions of text in the images introduce noise during the training process. Consequently, during inference, the model generates garbled and malformed text outside the target regions, which adversely affects the quality of the resulting images. To address this issue, a specific approach was implemented for handling small text in the poster dataset. This involved adding masks to the regions containing small text and utilizing the LaMa \cite{suvorov2022resolution} model to restore the images. Small text areas are text areas where the area obtained by PP-OCR accounts for less than 0.001 of the total area. The restored images are then incorporated into the poster dataset, ensuring improved quality.

To ensure high-resolution images in the dataset, we apply resolution filtering criteria, retaining only images with dimensions larger than 1024×1024 and a minimum shorter side of 768 pixels. To extract clean images containing glyphs from a vast amount of data, we implement filtering rules: (1) Only b-boxes with OCR recognition confidence greater than 0.8 for individual texts are retained. (2) Only text b-boxes with a character count of less than 15 are retained, and each image is limited to a maximum of ten b-boxes.(3) B-boxes whose center falls within 5\% of the image boundaries are excluded to eliminate the influence of bottom watermarks. (4) The center of each b-box is required to be at least 15\% away from the image boundaries in at least one direction.
(5) Bboxes with a single character area larger than 2000 square pixels are kept.

For the poster dataset, we employ refined processing techniques to ensure high quality. In addition to the aforementioned filtering criteria, aesthetic scoring is performed on the poster data, and LaMa restoration is applied to small glyph regions using added masks. 

\subsection{Dataset Statistics}
The statistics of our general dataset and poster dataset are shown below.

\begin{table}[hp]
\centering
\subtable[Chinese training data statistics.]{
\resizebox{1\linewidth}{!}{
\begin{tabular}{cccc}
\hline
Dataset         & \# Samples & \# Chars   & \# Unique Chars \\ \hline
General dataset & 1,365,425  & 25,412,521 & 5214  \\
Poster dataset  & 485,641  & 10,254,585 & 4585  \\ \hline
Total & 1,851,066  & 35,667,106 & 5299  \\ \hline
\end{tabular}}
\label{Table:chinese_table}
}
\qquad
\subtable[English training data statistics.]{        
\resizebox{1\linewidth}{!}{
\begin{tabular}{ccclc}
\hline
Dataset         & \# Samples & \# Chars   & \# Words  & \# Unique Chars \\ \hline
General Dataset & 1,195,221  & 9,343,222 & 1,832,111 & 245,865        \\ \hline
\end{tabular}}
\label{Table:english_table}
}
\caption{Statistical indicators of the dataset.}
\end{table}

\begin{figure}[ht]
	\centering
	\includegraphics[width=0.5\textwidth]{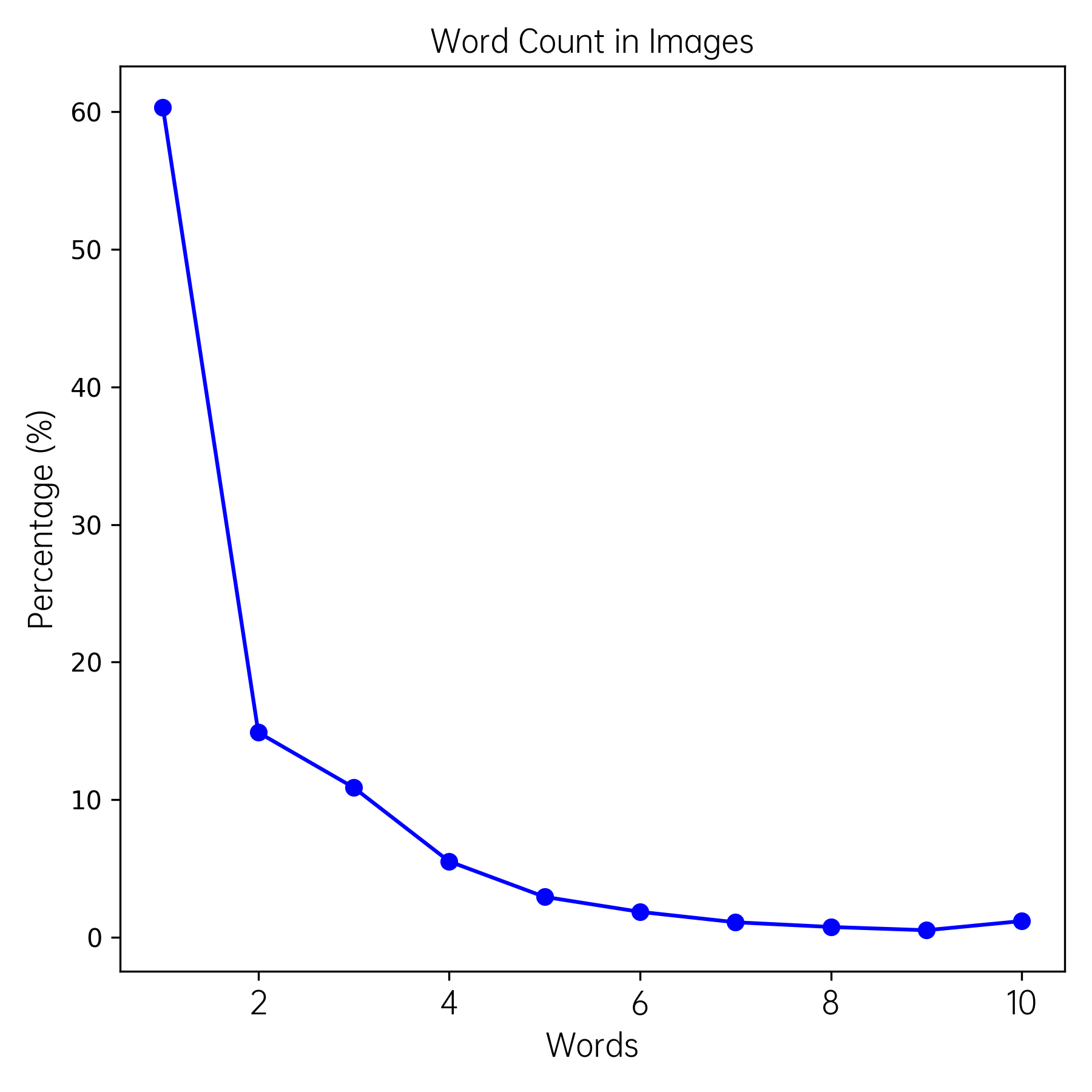}
 \caption{Statistical indicators of English data in general datasets.}
\label{fig:word_count}
\end{figure}
Particularly, the number of words per image in the general dataset was statistically analyzed, as shown in the Fig.\ref{fig:word_count}. The proportion of images with 1 word is the highest, at around 60\%. The percentage then gradually decreases as the number of words per image increases up to 10.

\begin{figure}
	\centering
	\subfigure[Line distribution of general dataset.]{
		\begin{minipage}[b]{0.3\textwidth}
			\includegraphics[width=1\textwidth]{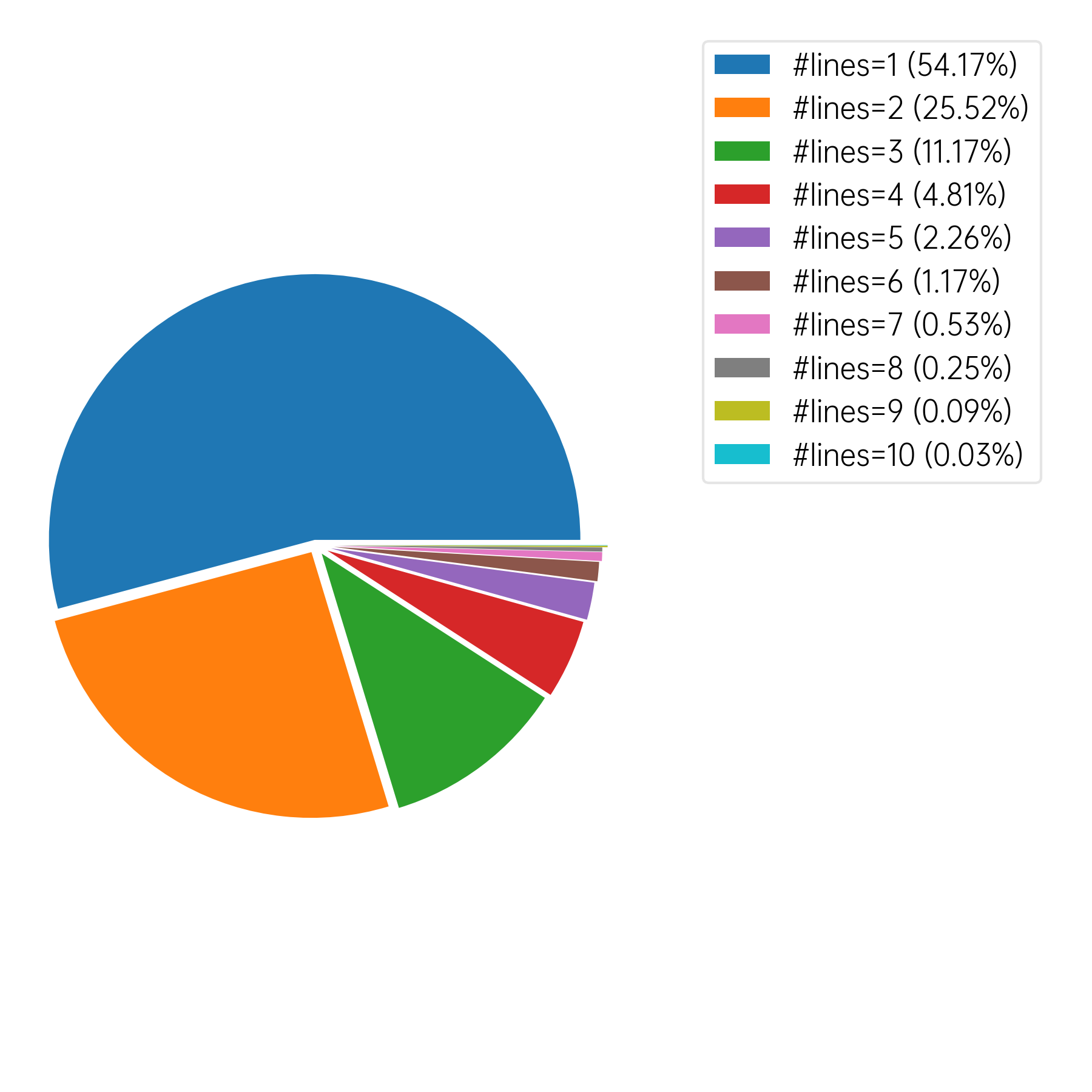}
		\end{minipage}
		\label{fig:bbox_general_dataset}
	}
    \\
    \subfigure[Line distribution of poster dataset.]{
    		\begin{minipage}[b]{0.3\textwidth}
   		 	\includegraphics[width=1\textwidth]{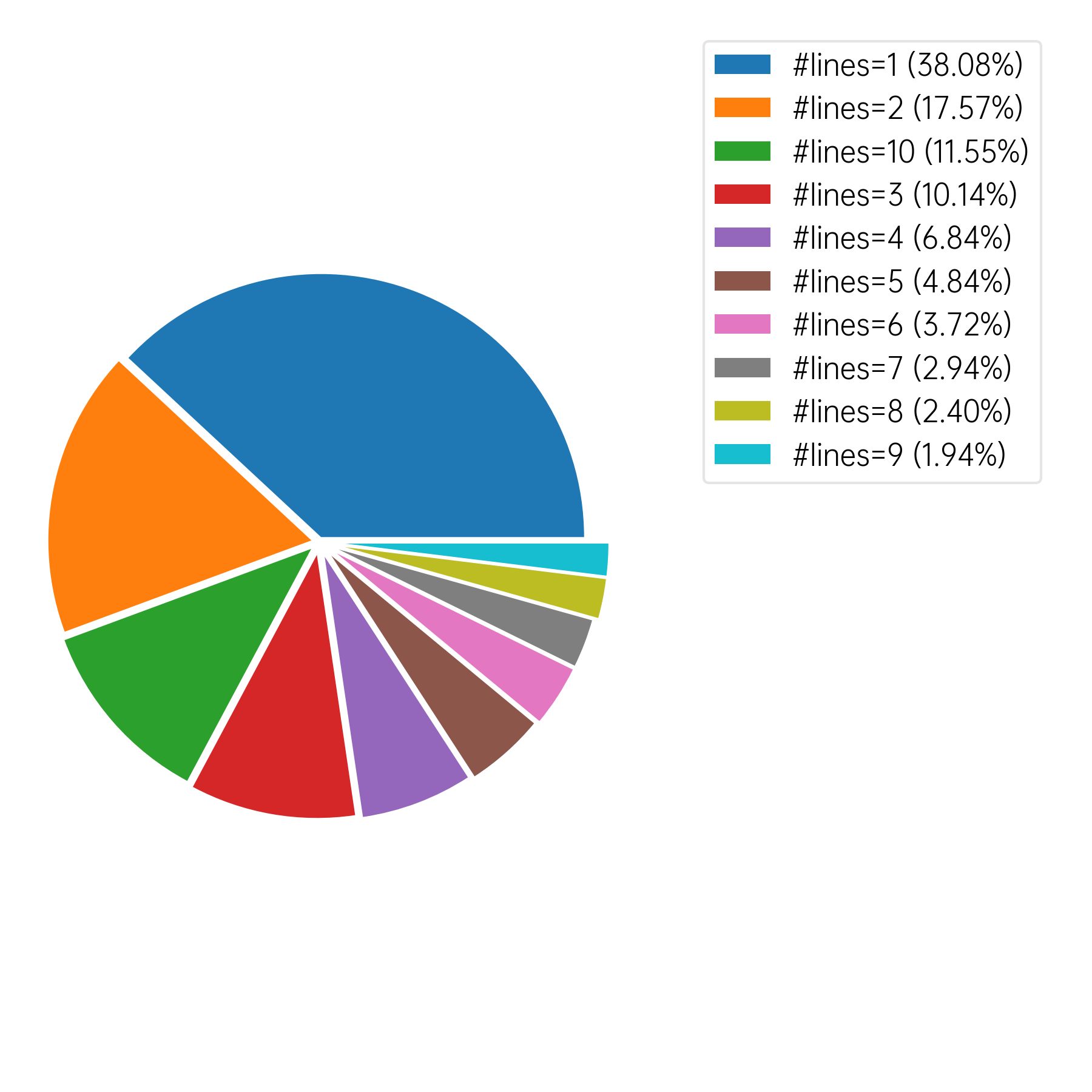}
    		\end{minipage}
		\label{fig:bbox_poster_dataset}
    	}
	\caption{Line distribution of GlyDraw2 dataset.}
	\label{fig:bbox_glydraw2_dataset}
\end{figure}

Fig.\ref{fig:bbox_glydraw2_dataset} analyzes the distribution of the number of text boxes in the general dataset and the poster dataset.
In the general dataset, the majority of images exhibit a concentration of 1-3 text boxes, with only a few containing 4-5 text boxes, and a relatively lower occurrence of 6 or more text boxes. In contrast, the poster dataset shows a more diverse distribution of text box numbers. Notably, approximately 11.55\% of the text boxes in this dataset have a count of 10, ranking third in proportion. Furthermore, the distribution of text boxes ranging from 5 to 9 shows a relatively balanced pattern. The poster dataset exhibits a richer diversity in the number of text boxes, which could be advantageous for training models to generate posters with more varied layouts.

\begin{figure}
	\centering
	\subfigure[Character count of general dataset.]{
		\begin{minipage}[b]{0.4\textwidth}
			\includegraphics[width=1\textwidth]{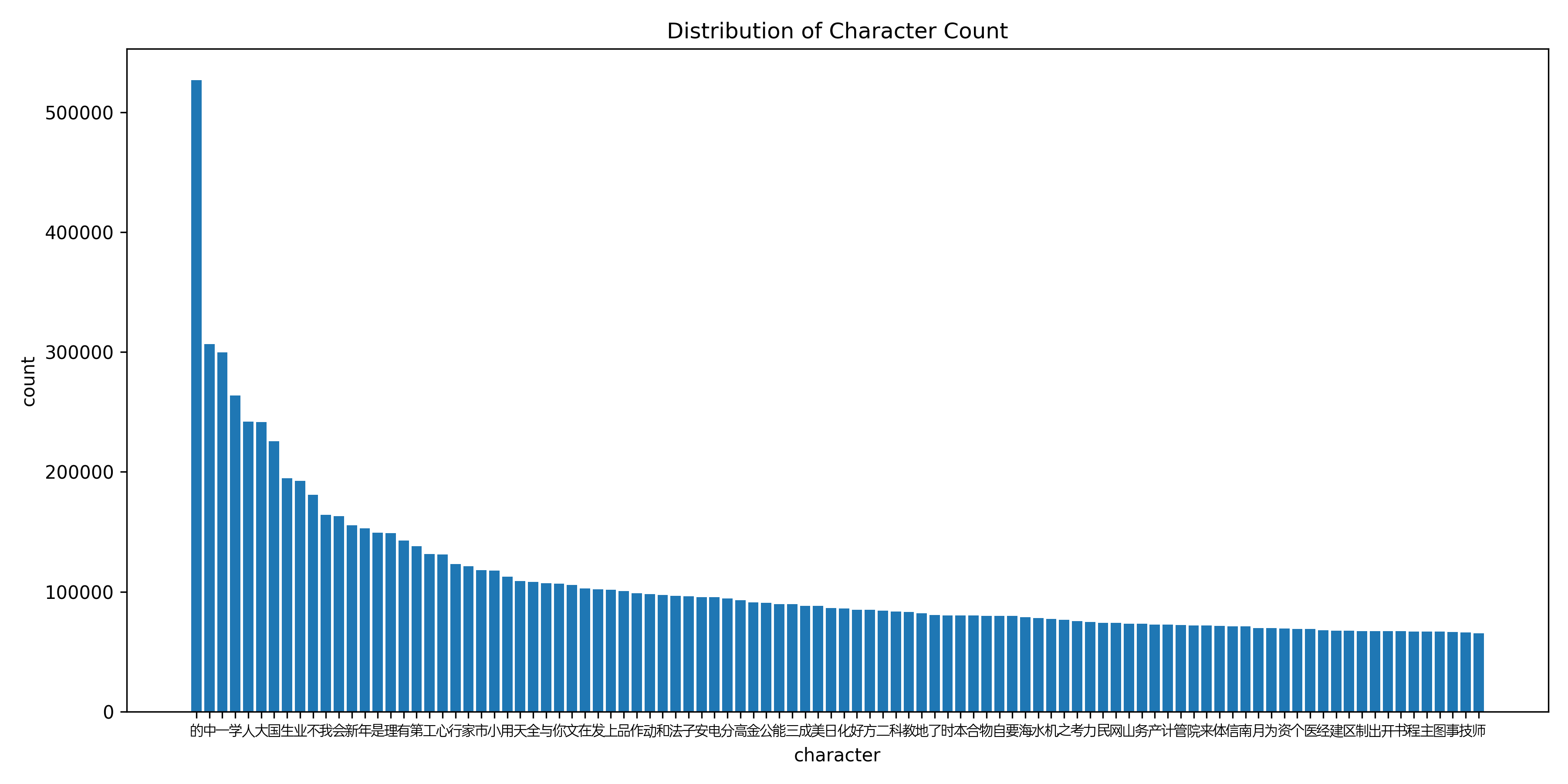}
		\end{minipage}
		\label{fig:character_general_dataset}
	}
    \\
    	\subfigure[Character count of poster dataset.]{
    		\begin{minipage}[b]{0.4\textwidth}
   		 	\includegraphics[width=1\textwidth]{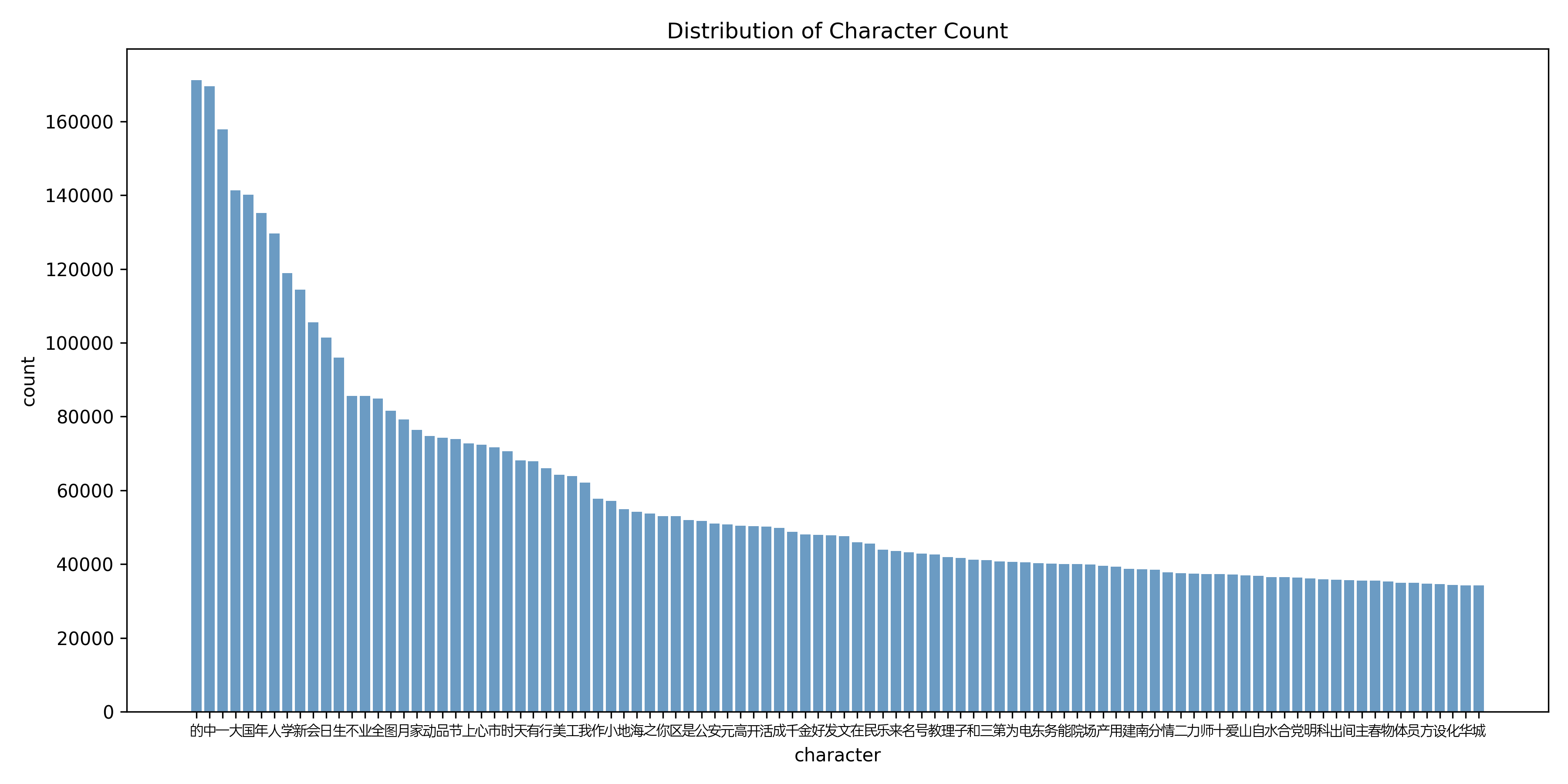}
    		\end{minipage}
		\label{fig:characte_poster_dataset}
    	}
	\caption{Character count of GlyDraw2 dataset.}
	\label{fig:characte_glydraw2_dataset}
\end{figure}

In addition, Fig.\ref{fig:characte_glydraw2_dataset} illustrates the 100 most frequent Chinese characters in the general dataset and the poster dataset. While the two datasets exhibit subtle differences in the most common Chinese characters, the poster dataset displays a more concentrated distribution of character frequencies, whereas the general dataset shows greater variance in character frequency.

\section{Implementation Details}
\label{Implementation}

The model we intend to train comprises two main components. The first component is a controllable T2I poster model, with the backbone of our framework being based on SDXL.
To adapt the multilingual understanding capacity of the SDXL encoder and maintain linguistic coherence between the prompt's description of the poster background and the generated text, we have incorporated the PEA-Diffusion strategy \cite{ma2023pea} into the backbone architecture. This strategy entails replacing the original SDXL encoder with a multilingual CLIP encoder and an adapter, followed by applying knowledge distillation to align semantic representations. 
Our model has a total of 1.6 billion trainable parameters, comprising the ControlNet and two additional cross-attention structures. 
Based on the characteristics of ControlNet and adapter, our solution has good portability.
We use the AdamW optimizer \cite{loshchilov2017decoupled} and set the learning rate to 3e-5. During the training phase, we adopt a two-stage progressive training strategy.
For initial training stage, the objective is to impart the model with text generation capabilities and the model was trained for 80,000 steps on the synthetic dataset without integrating the AAL for semantic consistency. Then in the second stage, a poster dataset with rich layouts was utilized. To maintain a diverse range of backgrounds in the generated posters, the model underwent training for 20,000 steps with AAL. The entire diffusion model is trained on 64 A100 GPUs for 10W steps with a batch size of 2 per GPU.

The second component is a layout generation model based on LLMs. 
We employed Baichuan2~\cite{yang2023baichuan2openlargescale} specifically for this task, using a training dataset consisting exclusively of poster data.This task requires predicting the content of the characters to be rendered in the entire text description, as well as the corresponding position coordinates for the character content, it posed a major challenge to the LLMs with a valid structured output. To improve prediction accuracy, we normalized the coordinate points and focused solely on utilizing only the top-left and bottom-right corner points.
In addition, to maintain the stability of the automatic generation process, a random rule-based layout generation approach was utilized when encountering invalid predictions from the LLMs. This involved integrating random strategies into the layout generation procedure. The implementation ratio of these random strategies was approximately 5\% to strike a balance between stability and variability in the generated layouts. The LLMs model for layout generation is trained on 64 A100 GPUs for 30K steps with a batch size of 10 per GPU.

\section{Details of Evaluation Benchmarks}
\label{Self-Constructed}

\textbf{Complex-Benchmark} consists of 200 prompts which include bilingual Chinese and English. In the Chinese prompts, the characters to be rendered are randomly combined and arranged, while the English prompts feature longer words with consecutive repetitions of letters. 
Specifically, for the Chinese language, we randomly combined characters from a pool of 2000 commonly used Chinese characters as the text to be rendered, resulting in a set of 100 prompts. The number of rows and characters per row were also randomly determined, ensuring the generation of prompts with a complete sense of randomness. The set of 100 prompts we devised comprises characters with intricate strokes and structures, such as ``\begin{CJK}{UTF8}{gbsn}薯\end{CJK}(potato)'', ``\begin{CJK}{UTF8}{gbsn}寨\end{CJK}(stockade)'', and ``\begin{CJK}{UTF8}{gbsn}聚\end{CJK}(gather)''. 
Although the number of evaluation samples is limited, they encompass a diverse range of frequently encountered Chinese characters, including some complex structural characters that are infrequently represented in the training dataset. Consequently, these prompts provide a robust means to holistically assess the model's Chinese character generation capability. For English text, we selected words with consecutive repeated letters and some longer words for rendering with 100 prompts. These words are prone to errors, making them persuasive indicators of the rendering proficiency for English words. Also, in contrast to AnyText-Benchmark, we provide a bbox that can render phrases and sentences, not just single words. This approach inevitably increases the difficulty of rendering.

\textbf{Poster-Benchmark} includes 240 prompts that describe the generation of posters.
To assess the automatic capabilities of our poster generation model, we specifically designed a dedicated dataset for poster evaluation, encompassing a variety of prompt forms for poster generation. This comprehensive dataset describe posters in both English and Chinese, enabling the generation of images in various resolutions, including landscape, portrait, and square formats. Unlike AnyText-Benchmark, which only allows English words inputs in text prompts, our model accommodates complete English sentences, thus facilitating the presentation of desired text.
Its purpose is to evaluate the layout accuracy, robustness, and overall aesthetic quality of automatic poster generation.

\section{GlyphDraw1.1}
\label{GlyphDraw1.1}

\begin{figure*}[ht!]
    \includegraphics[width=1\textwidth]{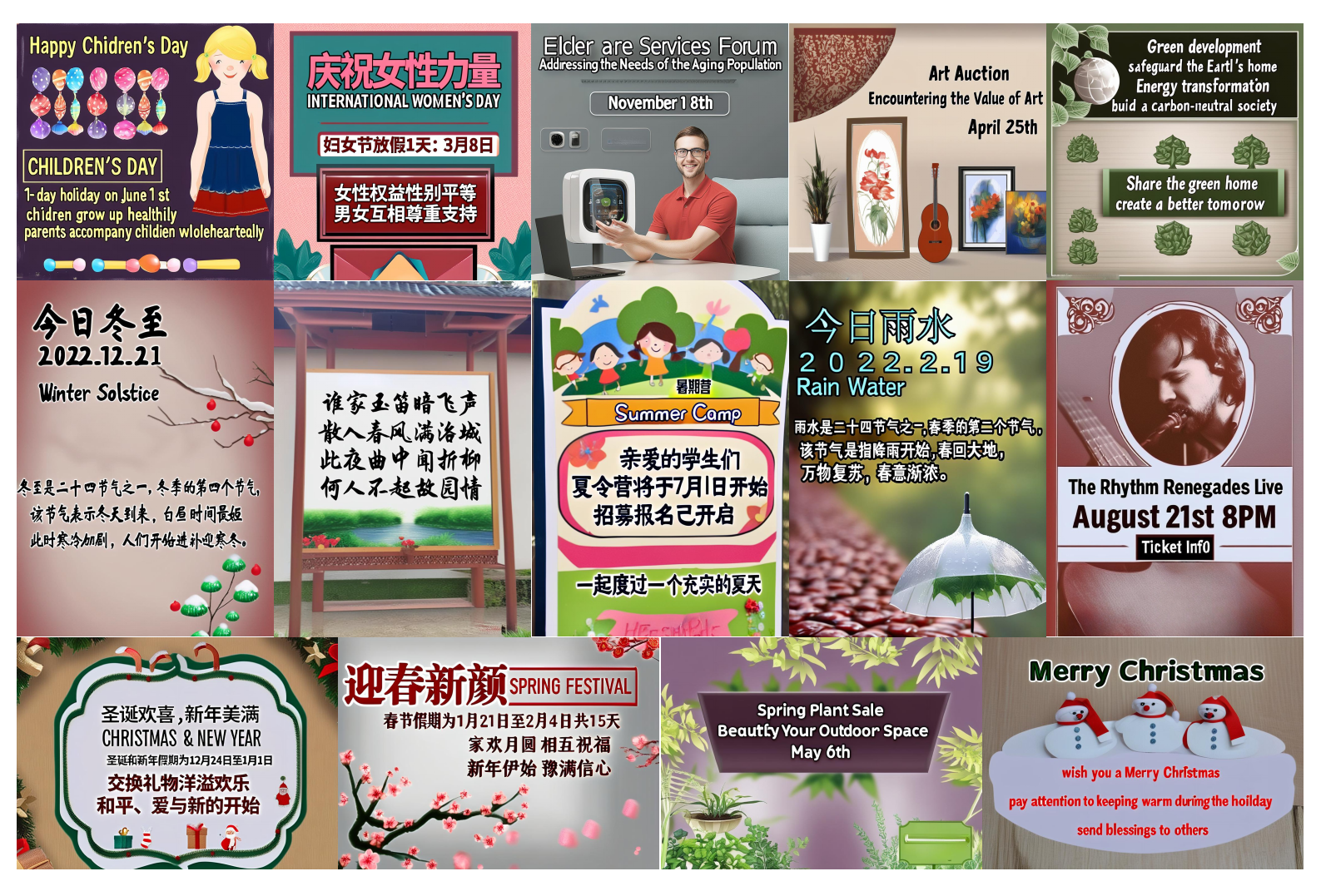}
     \caption{\small{Illustration of subjective results using GlyphDraw1.1}}
  \label{fig:banner}
\end{figure*}

Fig.\ref{fig:banner} demonstrates the raw image results under the GlyphDraw1.1 framework. In GlyphDraw1.1, the conditional input directly uses images rendered with fixed fonts instead of Canny images of actual fonts. From the displayed results, although good effects can be achieved for the background and small text, attributes such as font style and color cannot be controlled and are full of randomness. Moreover, this approach is difficult to fit when compared to GlyphDraw2, resulting in higher training difficulty and cost.

\subsection{Ablation Studies}
This ablation experiment was conducted in a subtractive manner. Given that we have numerous ablation experiments and wishing to reduce training costs, we uniformly set the first training phase in each experiment to 20,000 steps and the second phase to 10,000 steps and performed on the Chinese evaluation dataset.
The ablation studies involve examining 4 main aspects, namely: the impact of TCA and its specific modules; 2) the impact of AAL; 3) the impact of text encoder fusion; 4) the impact of ControlNet's condition input.

\begin{table}[ht]
\resizebox{1\linewidth}{!}{
\begin{tabular}{ccccc}
\hline
\multirow{2}{*}{Model} & \multicolumn{4}{c}{Chinese}   \\ \cline{2-5} 
   & Acc   & NED   & ClipScore       & HPSv2 \\ \hline
w/o CAG      & 0.7841& 0.8970& 0.4058& 0.2446\\
w/o CAC      & 0.7985& 0.9024& 0.3974& 0.2401\\
w/o TCA      & 0.7802& 0.8795& 0.3964& 0.2405\\
w/o AAL       & 0.8198& 0.9345& 0.3884& 0.2301\\
w/o FTE      & 0.7965& 0.9010& 0.4012& 0.2382\\
w/o CC       & 0.7845& 0.8975& 0.4001& 0.2422\\
\textbf{GlyphDraw1.1}& \textbf{0.8058} & \textbf{0.9125} & \textbf{0.3996} & \textbf{0.2412} \\ \hline
\end{tabular}}
\caption{Ablation Results on Poster-Benchmark in Chinese.}
\label{table2}
\end{table}

\textbf{TCA.}
TCA adds two CA layers, and here we individually ablate each added CA layer. Among them,  CAG represents the ablation of the CA interaction where the glyph feature as K, V is involved. Since the addition of this CA layer is intended to improve glyph accuracy, as shown in Table \ref{table2}, removing this layer results in a slight drop in accuracy but a certain improvement in the clip score and preference score. This indicates that while CAG improves the accuracy of text rendering, it sacrifices some text semantic alignment capability.
CAC represents the ablation of the adaptive CA interaction process that derives features from the ControlNet encoder. Here, both indicators will drop slightly, implying that the adaptive feature interaction can indeed enhance both the accuracy of text rendering and the ability for text semantic alignment as well as preference score.
TCA carries out the ablation of the entire TCA block. Similar to CAC both accuracy and preference score will decrease, further illustrating that the TCA module positively affects both text rendering accuracy and the preference score of the image.

\textbf{AAL.}
As seen in Table \ref{table2}, this strategy does indeed enhance the ability for semantic alignment and image quality to a certain degree, but it also sacrifices some text rendering accuracy. However, the overall impact is still positive.

\textbf{FTE.}
The primary purpose of the FTE is to ensure harmony between the font and the background. As can be observed from the ablation study Table \ref{table2}, all metrics are influenced to a certain extent. The FTE incorporates font feature information, which enhances the accuracy of text rendering. However, the fusion of image modalities may weaken the alignment of text semantics, leading to a slight decline in ClipScore. Lastly, the enhancement of image compatibility positively affects the preference score.

\textbf{ControlNet's condition.}
The condition input of ControlNet (CC) mainly affects the accuracy of the glyph, reducing the influence of the descriptive caption of the image on text rendering and to some extent improving glyph accuracy.

\section{LLMs Layout Prediction Experiment.} 
\label{LLMs}

Firstly, we constructed four tasks according to the difficulty level.

\begin{itemize}
  \item [1.] 
Input: Caption describing the image containing the glyph to be rendered and the size of the image to be generated; 
Output: The glyph to be rendered and the four coordinate points of the corresponding bbox, with multiple similar tuples corresponding to multiple positions.
  \item [2.] 
Input: Caption describing the image containing the glyph to be rendered; 
Output: The glyph to be rendered and the four normalized coordinate points of the corresponding bbox, with multiple similar tuples corresponding to multiple positions.
  \item [3.] 
Input: Caption describing the image containing the glyph to be rendered and the size of the image to be generated; Output: The glyph to be rendered and the two coordinate points (top left and bottom right) of the corresponding bbox, with multiple similar tuples corresponding to multiple positions.
  \item [4.] 
Input: Caption describing the image containing the glyph to be rendered; Output: The glyph to be rendered and the two normalized coordinate points (top left and bottom right) of the corresponding bbox, with multiple similar tuples corresponding to multiple positions.
\end{itemize}

The first two tasks require predicting four position coordinates, which is the most challenging but meets the requirements the most. Normalization reduces the task difficulty but sacrifices some diversity to a certain extent by reducing the solving range. The last two tasks lower the fine-tuning difficulty, but similarly sacrifice the diversity of the predicted coordinates, meaning the bbox coordinates limit it to be a rectangle.

\begin{figure*}[tb!]
	\centering
	\includegraphics[width=1\textwidth]{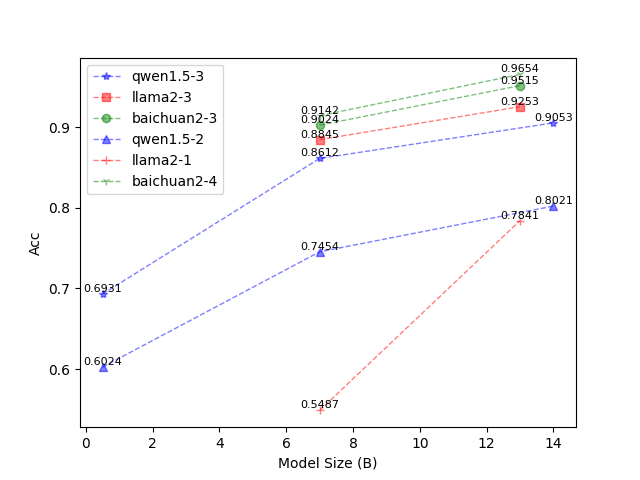}
 \caption{\small{Results of different LLMs for different data modes.}}
\label{llm}
\end{figure*}

The experimental results, shown in Fig.~\ref{llm}, the numerical suffix in the model name represents the task mode id. The experiment first discovered that the larger the model parameter volume, the better the fine-tuning effect. The results of output normalization have a higher accuracy rate. In the end, we chose the Baichuan2-13B model, with the third task mode.

\begin{figure*}[htbp]
	\centering
	\includegraphics[width=1\textwidth]{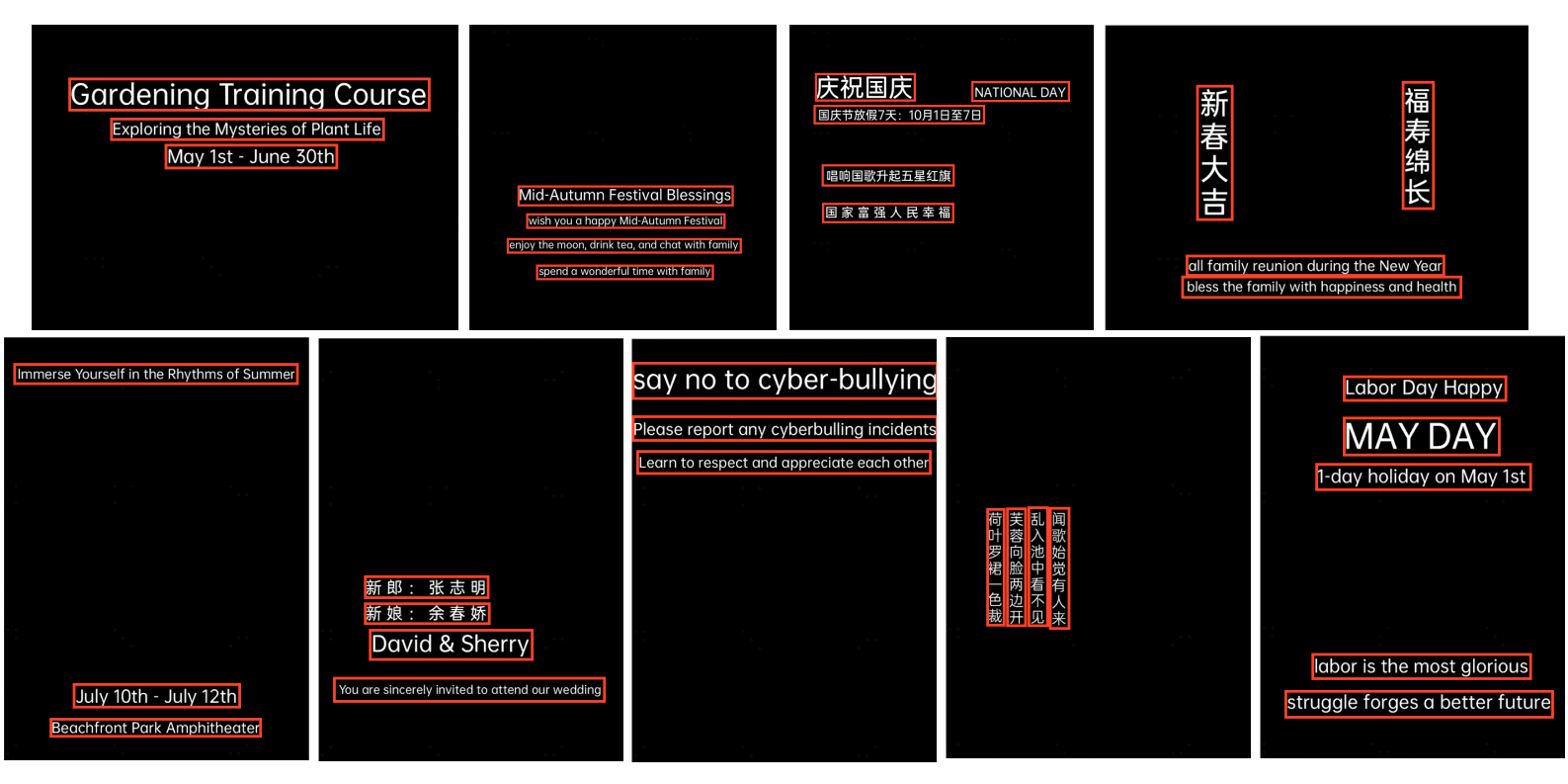}
 \caption{\small{Test the results of LLMs prediction in Poster-Benchmark.}}
\label{bbox}
\end{figure*}

Fig.~\ref{bbox} shows the results after fine-tuning the LLMs on our custom evaluation set. The main advantages are seen in three aspects. Firstly, in terms of the poster's title, the model tends to predict a b-box with a relatively large area. Secondly, the continuity of content in adjacent b-boxes offers contextual meaning, allowing the model to learn the semantic information required to render the glyph. Lastly, the size of the b-boxes tends to be proportional to the number of characters or words they contain.

\section{Multi-Style Font Generation}
\label{Style}

\begin{figure*}[htbp]
	\centering
	\includegraphics[width=1\textwidth]{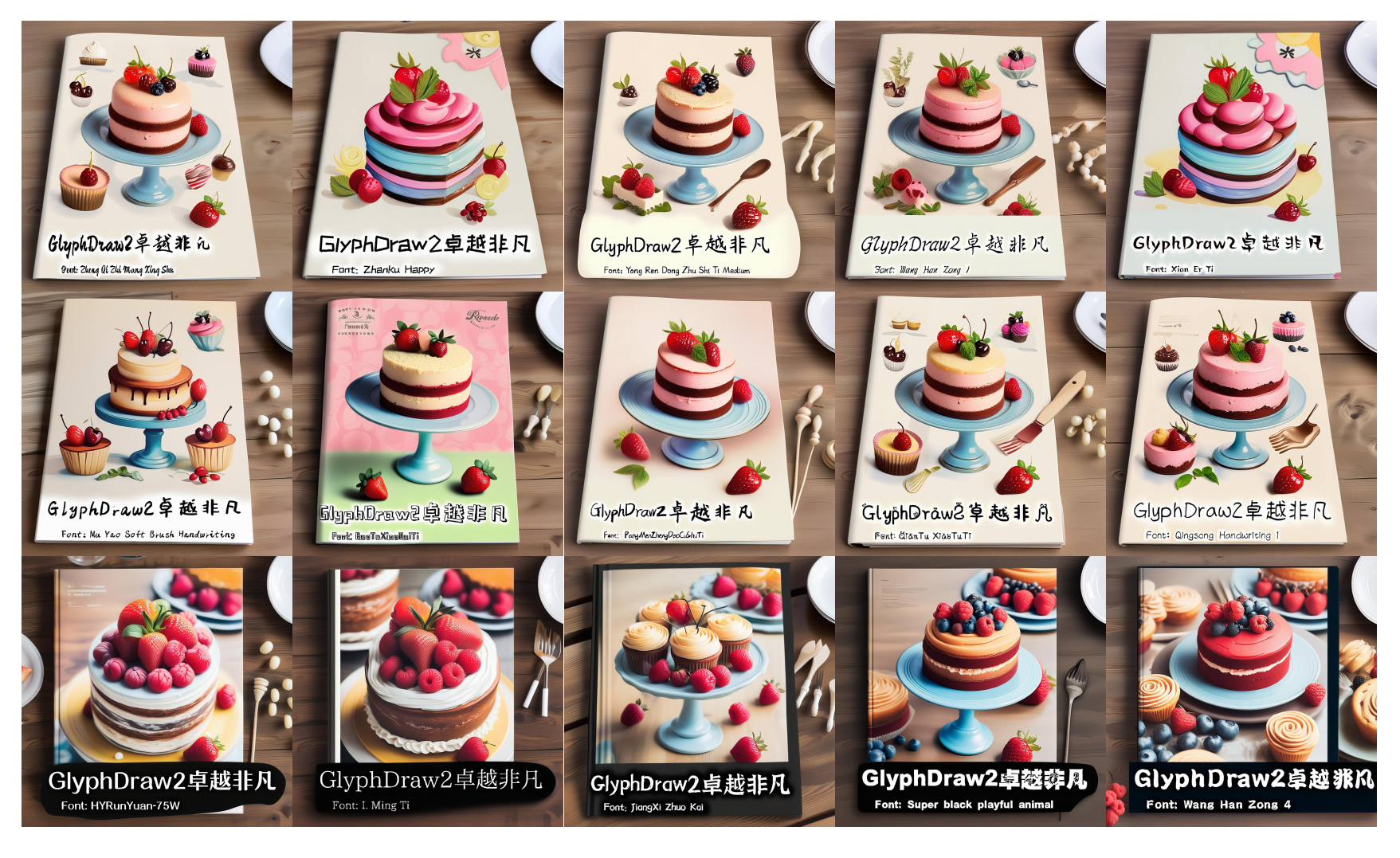}
 \caption{\small{Generated images in different fonts.}}
\label{Multi-Style}
\end{figure*}

In Fig.\ref{Multi-Style}, we present the controllable generation results for 15 different fonts, including both Chinese and English. prompts:"There is a book on the table, the content of which is various fruit cakes, with the title 'GlyphDraw2: Extraordinary Excellence', 'Font:***'. '***' represents a specific font type.

\section{Subjective Results Comparison}
\label{subjective}
To offer a more intuitive comparison of rendering effects, we display the results of various methods from Table 1 in the paper, featuring the three major text-to-image models (SD3, Kolors, Flux) along with open-source methods like TextDiffuser-2, AnyText, UDiffText, and Glyph-ByT5 for comparative analysis. 

Initially, we present the raw image prompts, "Winter scenery, snow falling, thousands of pear blossoms, Frost-kissed mornings," with the text to be rendered as "Frost-kissed mornings", "The snow falls as if the vernal breeze had come back overnight", "adorning thousands of pear trees with blossoms white.", and ``\begin{CJK}{UTF8}{gbsn}忽如一夜春风来,千树万树梨花开\end{CJK}(Ancient Chinese poetry, meaning a sudden night breeze of spring arrives, and thousands of pear trees blossom)'' comprising three lines in English and one in Chinese. 

\begin{figure*}[ht!]
	\centering
	\includegraphics[width=1\textwidth]{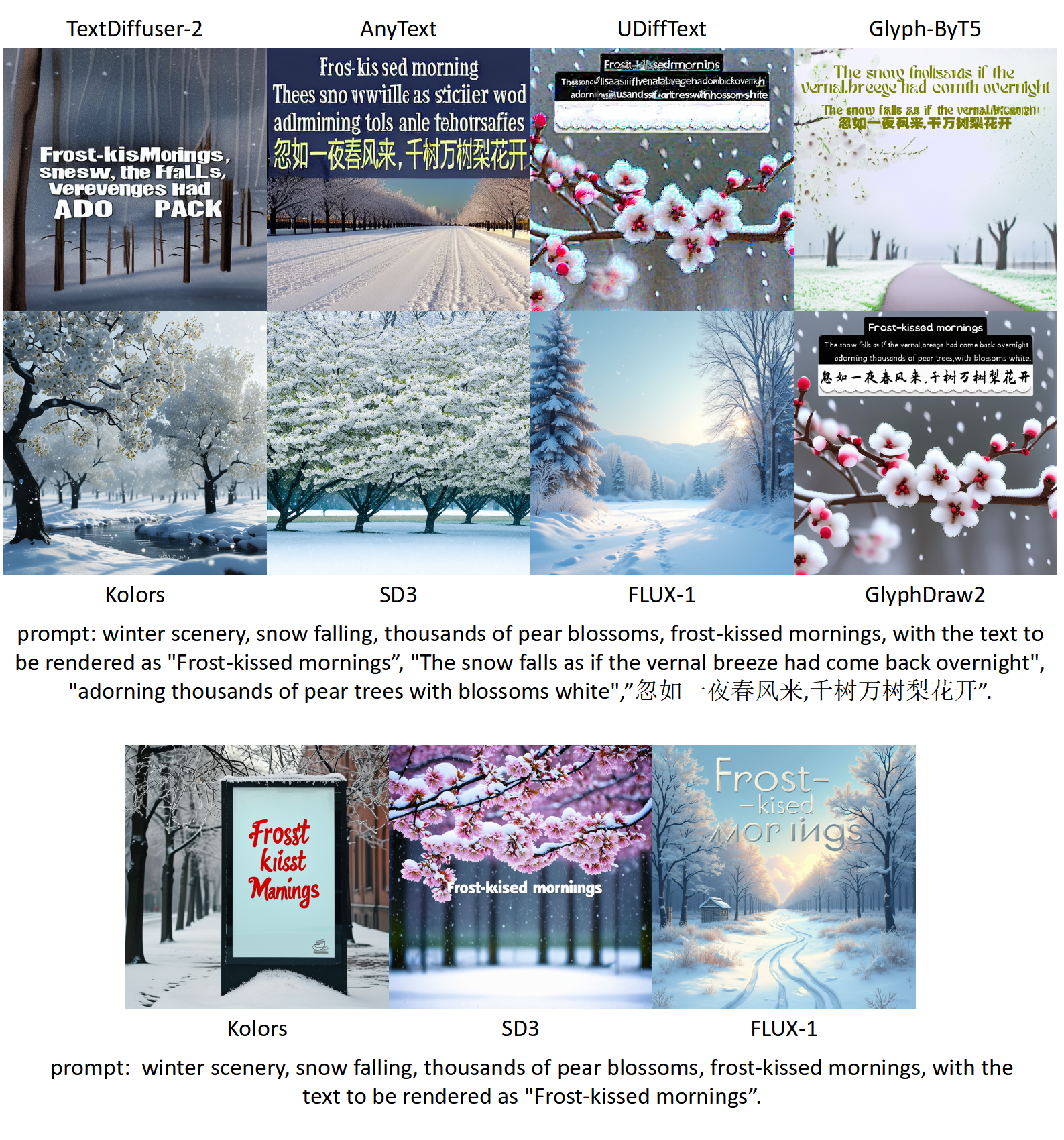}
 \caption{\small{Comparison of subjective results from various methods.}}
\label{contrast}
\end{figure*}

In the top row of Fig.\ref{contrast}, the four images from left to right are TextDiffuser-2, AnyText, UDiffText, and Glyph-ByT5, respectively. TextDiffuser-2 does not support Chinese, and its accuracy falls short when rendering lengthy text. For AnyText, we purposely enlarged the bbox to enhance the generation quality; however, the rendering effect for long text remains subpar. UDiffText's currently open-source model lacks text-to-image capability, so images generated by glyphdraw2 were used for editing and generation. The paper's experimental section mentioned that UDiffText can only generate up to 12 characters and does not support multiple text positions simultaneously. Therefore, we performed text segmentation and multiple editing generations, as UDiffText also does not support Chinese, the problem here is that after more than 5 iterations, which in our case was 23 iterations, the background starts to deteriorate significantly.
Then there's Glyph-ByT5, which is highly accurate in generating Chinese characters but struggles with long English sentences. Here we matched the b-box input of GlyphDraw2, but Glyph-ByT5 still produces renderings that sometimes exceed the b-box boundaries. 

The middle four images, from left to right, illustrate the generation effects of Kolors, SD3, FLUX.1-dev, and GlyphDraw2. These three large models are unable to render long text, even with multiple commands to enhance this capability. 

In the bottom three images, we only render "Frost-kissed mornings" for the three major models, with all showing good results. The most aesthetically appealing model is still the 12B FLUX.1-dev. Hence, for rendering long text, large models still have significant advancements to achieve in direct end-to-end generation.

\section{Summary of Text Rendering}
\label{related}
Table\ref{table_related} provides a detailed summary of font rendering work, categorizing it into four types according to the core problems solved by different work. The related work chapter of the paper also introduces the detailed content, and here we summarize it into a table for easy vertical comparison. Although the starting points and specific plans of each paper vary widely, Glyph-ByT5 and GlyphDraw2 are still very competitive in terms of the final presentation effect.

\begin{table*}[ht]
\resizebox{1\linewidth}{!}{
\begin{tabular}{ccc}
\hline
\textbf{Group}                              & \textbf{Methods}                 & \textbf{Key Features}      \\ \hline
\multirow{5}{*}{\textbf{\begin{tabular}[c]{@{}c@{}}Text Rendering Accuracy \\ \& Background Coherence\end{tabular}}} & GlyphDraw\cite{ma2023glyphdraw}   & Fuses font and text features into a diffusion model     \\
& TextDiffuser\cite{chen2023textdiffuser}& Adds Layout Generation module and Character-aware Loss  \\
& GlyphControl\cite{yang2024glyphcontrol}& Uses ControlNet\cite{zavadski2023controlnet} for Text Rendering \\
& AnyText\cite{tuo2023anytext}     & Incorporates auxiliary conditions like text glyph, position, masked image, and text perceptual loss \\
& Brush Your Text\cite{zhang2023brushtextsynthesizescene}                  & Proposes local attention constraint in cross-attention layer                                        \\ \hline
\multirow{4}{*}{\textbf{\begin{tabular}[c]{@{}c@{}}Character-Aware \\ Text Encoders\end{tabular}}}                   & UDiffText\cite{zhao2023udifftext}   & Lightweight character-level text encoder replacing CLIP encoder                                     \\
& Glyph-ByT5\cite{liu2024glyph}  & Fine-tunes a character-aware ByT5 encoder aligned with glyph features                               \\
& DreamText\cite{wang2024highfidelityscenetext}   & Jointly trains text encoders and generators             \\
& SceneTextGen\cite{zhangli2024layoutagnosticscenetextimage}& Employs character-level encoder to extract detailed character-specific features                     \\ \hline
\multirow{5}{*}{\textbf{\begin{tabular}[c]{@{}c@{}}Text Layout, Color, \\ \& High-Level Attributes\end{tabular}}}    & TextDiffuser-2\cite{chen2024textdiffuser}                   & \multirow{2}{*}{Uses LLMs to predict font layout}       \\
& ARTIST\cite{zhang2024artistimprovinggenerationtextrich}      &                            \\
& \begin{tabular}[c]{@{}c@{}}Refining Text-to-Image\cite{lakhanpal2024refiningtexttoimagegenerationaccurate} \end{tabular} & Employed text layout generator     \\
& Glyph-ByT5\cite{liu2024glyph}  & Incorporates font type and color control                \\
& CustomText\cite{paliwal2024customtextcustomizedtextualimage}  & Considers a variety of text attribute controls          \\ \hline
\multirow{3}{*}{\textbf{Data-Based T2I}}    & SD3\cite{esser2024scalingrectifiedflowtransformers} & \multirow{3}{*}{Strong image coherence with constrained character accuracy and number}              \\
& Kolors\cite{kolors}     &                            \\
& FLUX&                            \\ \hline
\end{tabular}}
\caption{Text rendering work summary.}
\label{table_related}
\end{table*}

\section{Limitations and Future Outlook}
\label{Limitations_a}
Firstly, for the glyph bboxes predicted by LLMs, the prediction accuracy is meager for complex scenarios, such as when a user inputs a paragraph of text without quotation marks as a bbox prompt. Secondly, balancing the richness of background generation and the accuracy of text rendering is still relatively difficult. In our current approach, we prioritize glyph accuracy; thus, the visual appeal of the background may be weaker. Additionally, the generation accuracy for tiny glyphs or paragraph texts still needs improvement. Fig.\ref{limitation} shows some failed examples, mainly in three aspects:

\begin{figure*}[ht]
	\centering
	\includegraphics[width=1\textwidth]{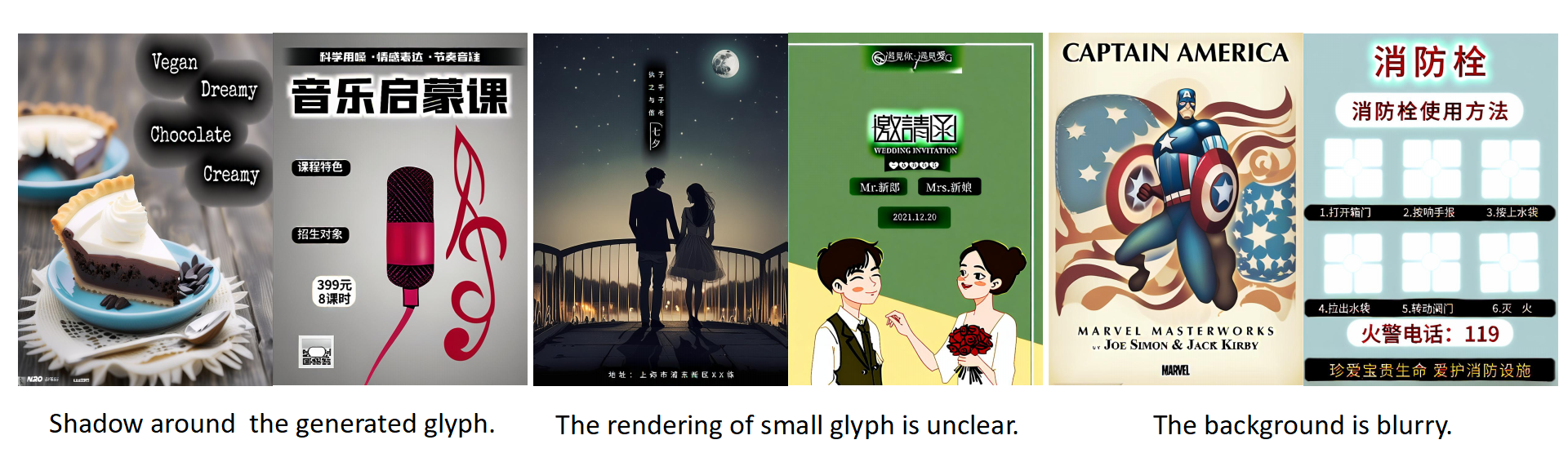}
 \caption{\small{Failed examples.}}
\label{limitation}
\end{figure*}

\begin{itemize}
  \item [1.] 
    \textbf{Shadow around the generated glyph.} When we tested posters containing more characters, we found that shadows appeared around the rendered text. Upon further investigation, we discovered that this issue originates from the noise in the edge detection of text by the conventional Canny algorithm when dealing with complex background images. This noise directly affects the training results. In the future, we plan to abandon the conventional Canny calculation and use a specially optimized model\cite{yu2024eaformerscenetextsegmentation} for edge detection.
  \item [2.] 

\textbf{The rendering of small glyphs is unclear.} As shown in the middle two pictures of Fig.\ref{limitation}, when the size of the rendered text is reduced to a certain threshold, the rendering accuracy declines significantly. The main reason for this issue is the large errors in the edge detection of small characters by Canny. Improvements can be made by optimizing the Canny.
  \item [3.] 

\textbf{The background is blurry.} The overall richness of the background is affected by two factors, as shown in the two pictures on the right side of Fig.\ref{limitation}. On one hand, it is easily influenced by the prompt; on the other hand, the layout of the characters plays a major role.
\end{itemize}

\textbf{ Future Outlook.}
In the future, our research direction mainly includes two paths. The first is to further explore along the framework of MMDiT. MMDiT uses a combination of DiT Block with adaLN-Zero and an improved version of DiT Block with In-Context Conditioning in the interaction of text. Unlike conventional cross-attention, which only introduces condition information at a certain layer of the block, the information interaction of MMDiT can pay more attention to some details as the network layer deepens. The second is to comprehensively combine the semantic information of the text to be rendered and the image encoding information, and then deeply explore the interaction of the condition and image token information with MMDiT.

\clearpage



\end{document}